\documentclass[sigconf]{acmart}
\pdfoutput=1

\copyrightyear{2022} 
\acmYear{2022} 
\setcopyright{acmcopyright}\acmConference[ICCAD '22]{IEEE/ACM International Conference on Computer-Aided Design}{October 30-November 3, 2022}{San Diego, CA, USA}
\acmBooktitle{IEEE/ACM International Conference on Computer-Aided Design (ICCAD '22), October 30-November 3, 2022, San Diego, CA, USA}
\acmPrice{15.00}
\acmDOI{10.1145/3508352.3549397}
\acmISBN{978-1-4503-9217-4/22/10}

\usepackage{xspace}
\usepackage{xcolor}
\usepackage{multirow}
\usepackage{multicol}

\newcommand\ssr{Smart Scissor\xspace}
\newcommand\ssd{SS-DIC\xspace}
\newcommand\css{SS-DIC-CS\xspace}
\newcommand\cs{SS-CS\xspace}
\newcommand\rcc{RCC\xspace}
\newcommand\sota{SOTA\xspace}

\newcommand*{\affaddr}[1]{#1} % No op here. Customize it for different styles.
\newcommand*{\affmark}[1][*]{\textsuperscript{#1}}

\begin{document}

% \title{Smart Scissor: Reducing Spatial Redundancy of CNNs for Embedded Hardware via Dynamic Image Cropping}
\title{Smart Scissor: Coupling Spatial Redundancy Reduction and CNN Compression for Embedded Hardware}

% \author{Hao Kong}
% % \authornote{Both authors contributed equally to this research.}
% \email{kong.hao@ntu.edu.sg}
% % \orcid{1234-5678-9012}
% % \author{G.K.M. Tobin}
% % \email{webmaster@marysville-ohio.com}
% \affiliation{%
%   \institution{HP-NTU Digital Manufacturing Corporate Lab, Nanyang Technological University}
% %   \streetaddress{P.O. Box 1212}
% %   \city{}
% %   \state{Ohio}
%   \country{Singapore}

\author{\Large
Hao Kong\affmark[1,2], Di Liu\affmark[2], Shuo Huai\affmark[1,2], Xiangzhong Luo\affmark[1], Weichen Liu\affmark[1], \\
Ravi Subramaniam\affmark[3], Christian Makaya\affmark[3], and Qian Lin\affmark[3]\\
\affaddr{\normalsize \affmark[1]School of Computer Science and Engineering, Nanyang Technological University, Singapore}\\
\affaddr{\normalsize \affmark[2]HP-NTU Digital Manufacturing Corporate Lab, Nanyang Technological University, Singapore}\\
\affaddr{\normalsize \affmark[3]HP Inc., Palo Alto, California, USA}
% \affaddr{\LaTeX\ University}%
}


\begin{abstract}
  Scaling down the resolution of input images can greatly reduce the computational overhead of convolutional neural networks (CNNs), which is promising for edge AI. However, as an image usually contains much spatial redundancy, e.g., background pixels, directly shrinking the whole image will lose important features of the foreground object and lead to severe accuracy degradation. In this paper, we propose a dynamic image cropping framework to reduce the spatial redundancy by accurately cropping the foreground object from images. To achieve the instance-aware fine cropping, we introduce a lightweight foreground predictor to efficiently localize and crop the foreground of an image. The finely cropped images can be correctly recognized even at a small resolution. Meanwhile, computational redundancy also exists in CNN architectures. To pursue higher execution efficiency on resource-constrained embedded devices, we also propose a compound shrinking strategy to coordinately compress the three dimensions (depth, width, resolution) of CNNs. Eventually, we seamlessly combine the proposed dynamic image cropping and compound shrinking into a unified compression framework, \ssr, which is expected to significantly reduce the computational overhead of CNNs while still maintaining high accuracy. Experiments on ImageNet-1K demonstrate that our method reduces the computational cost of ResNet50 by 41.5\% while improving the top-1 accuracy by 0.3\%. Moreover, compared to HRank, the state-of-the-art CNN compression framework, our method achieves 4.1\% higher top-1 accuracy at the same computational cost. The codes and data are available at \textcolor{blue}{\url{https://github.com/ntuliuteam/smart-scissor}}
  
%   At inference stage, we first use the box predictor to predict the bounding box of the foreground objective and crop the objective for inference. In such way, we can use a smaller resolution to reduce the computational overhead while still maintain a high accuracy. Moreover, as the resolution gets smaller, the accuracy drop gets bigger, we propose to jointly shrinking multiple dimensions to obtain a higher model compression ratio and a better trade-off between model overhead and accuracy.
\end{abstract}

\maketitle
\pagestyle{plain}
\pagenumbering{gobble}

\section{Introduction}
\label{sec:intro}

Modern convolutional neural networks (CNNs) continue to break previous accuracy records with the advances in large-scale datasets \cite{deng2009imagenet,lin2014microsoft,cordts2016cityscapes, ridnik2021imagenet} and network architecture innovation \cite{tan2019efficientnet,liu2018darts,wu2019fbnet,liu2022convnet}. Naturally, the accuracy improvement comes at the cost of higher computational overhead \cite{tan2019efficientnet,brown2020language,fedus2022switch}. Recently, there is a trend deploying CNNs in edge environments \cite{shi2016edge} to mitigate the latency and privacy concerns. However, the prohibitive computational cost impedes the deployment of advanced models onto resource-constrained edge devices. To enable edge applications like autopilot to benefit from the development of neural networks, efforts have been made to compress CNNs for a better trade-off between accuracy and model overhead, such that they can be easily deployed onto various edge hardware devices. 

\begin{figure}
    \centering
    \includegraphics[width=0.45\textwidth]{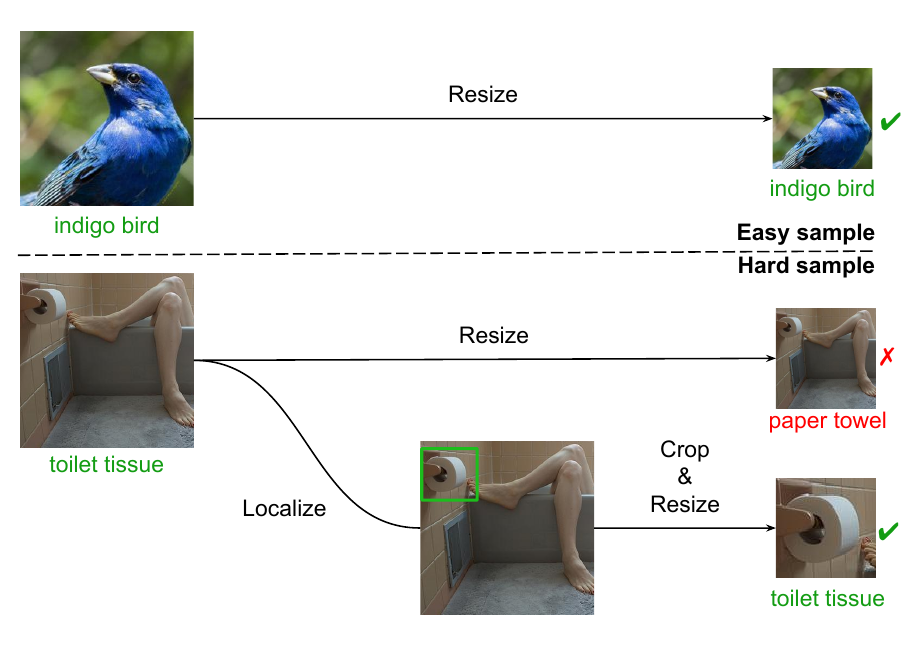}
    \caption{The prediction results of our pretrained ResNet-50 model. For easy samples, the network can still generate correct predictions at a smaller resolution (e.g. 112 $\times$ 112 for ImageNet). For hard samples, simply resizing the images to a smaller resolution can lead to misclassification, while the dynamic cropping strategy can correctly classify hard samples at a smaller resolution.}
    \label{fig:motivation}
\end{figure}

The computational cost, i.e., the Multiply-Accumulate Operations (MACs) of a CNN mainly result from gigantic network architectures and high-resolution input images (e.g., 224$\times$224 for ImageNet). To reduce the cost, on the one hand, network architecture pruning \cite{lin2020hrank,wang2021convolutional,yu2019slimmable,gao2021network} devotes to removing the redundancy in network architectures to construct more compact networks. For instance, MobileNets \cite{sandler2018mobilenetv2} and Slimmable networks \cite{yu2019slimmable} yield `thinner' networks by using less channels in each layer, which reduces the computation and memory footprint. On the other hand, as the MACs of a CNN reduce quadratically with respect to the image resolution, many works resize the input images to a smaller resolution (e.g., 112$\times$112) to reduce the computational cost \cite{sandler2018mobilenetv2,tan2019mnasnet,zhu2021dynamic}. However, different images correspond to diverse classification difficulties. As demonstrated in Figure \ref{fig:motivation}, easy samples with clear foreground can be correctly recognized even at a smaller resolution. While for hard samples, as the foreground object only occupies a small portion of the whole image, directly shrinking the image will lose the details of the object, leading to a wrong prediction. In contrast, if we only crop the foreground patch for inference, hard samples can also be correctly classified at a small resolution. However, existing image preprocessing methods, e.g., ResizedCenterCrop (RCC), crop all images in a static manner and cannot achieve such instance-aware fine cropping, which motivates our work.

In this paper, we propose a dynamic image cropping (DIC) framework to facilitate the inference with low-resolution images, thereby reducing the costs of CNNs and boosting the deployment of CNNs onto resource-constrained edge devices. 
DIC first efficiently localizes the most discriminative foreground of the input image with a lightweight foreground predictor, then the detected foreground region will be preserved and the redundant background will be discarded. By this means, DIC is capable of generating fine-cropped images with less spatial redundancy, thereby improving the inference accuracy even under low-resolution settings. To train the foreground predictor on large-scale classification datasets, we first employ a visualization technique, Grad-CAM \cite{selvaraju2017grad}, to generate the salience map of each training image to indicate the most discriminative foreground. Based on the salience map, we then automatically generate a bounding box annotation for each training image. Subsequently, we use the image-box pairs to train the predictor in a supervised manner. Once the foreground predictor is trained, it can be directly applied to any CNN backbones without modification.

Meanwhile, to deal with the redundancy in network architectures for higher execution efficiency, we propose a compound shrinking (CS) strategy to jointly compress the three dimensions (depth, width, resolution) of a CNN. We first investigate the impact of shrinking each dimension on accuracy. Based on that, we then calculate the optimal shrinking coefficient for each dimension to coordinate the shrinking of different dimensions to achieve the highest compression rate while still maintaining the accuracy. Eventually, DIC and CS are seamlessly combined to form a deep compression framework, \ssr, where an image is firstly cropped by DIC, then the cropped patch will be resized to the resolution calculated by CS and sent to the model compressed by CS for efficient inference. With our approach, both the spatial redundancy in images and the architecture redundancy in networks are reduced and thus the execution of CNNs on embedded hardware is accelerated.

The main contributions of this paper are summarized as follows:
\begin{enumerate}
    % \item We propose a dynamic image cropping framework to reduce the computational overhead of CNNs brought by high-resolution images. By automatically removing the redundant background in images and generating finely cropped images, our framework can yield accurate predictions with a smaller resolution, which significantly reduces the computational cost and improves the execution efficiency on hardware.
    \item We propose a dynamic image cropping framework to reduce the spatial redundancy in images. We introduce a lightweight foreground predictor to efficiently localize the foreground object and conduct instance-aware dynamic cropping. With the finely cropped images, CNNs can yield accurate predictions using a smaller resolution, which greatly reduces the computational cost of CNNs.
    
    % \item We design an accurate predictor to efficiently predict the foreground of an image. The predictor is well designed with negligible parameters and computation, which can quickly determine the foreground area and consequently accelerate the whole pipeline of classification.
    
    \item We also propose a compound shrinking strategy to coordinately shrink the three dimensions (depth, width, resolution) of a CNN. We first quantify the impact of each dimension on accuracy, and then compute the optimal shrinking coefficient for each dimension accordingly. By this means, we can greatly reduce the redundancy in network architectures while still maintaining high accuracy.

    \item We seamlessly combine the dynamic image cropping and compound shrinking into a deep compression framework, which is able to optimally adapt the model cost to meet different resource constraints of embedded hardware.
    
\end{enumerate}

Experiments on ImageNet-1K demonstrate that our method reduces the MACs of ResNet50 by 41.5\% while improving the top-1 accuracy by 0.3\%. In addition, our method achieves 4.2\% higher top-1 accuracy with similar MACs and 54.3\% less parameters compared to the most widely used image cropping method, ResizedCenterCrop (RCC). In low MACs regime, our approach also outperforms HRank \cite{lin2020hrank} by a remarkable improvement (4.1\%) in top-1 accuracy.

\section{Related Work}
\label{sec:related}
\ssr is mainly related to object discovery and neural network compression, so we discuss the related works in this section. 
\noindent
\textbf{Object discovery: }
Object discovery algorithms have made impressive progress. Among them supervised object detection (SOD) \cite{bochkovskiy2020yolov4,liu2016ssd,ren2015faster,he2017mask} achieves relatively high accuracy on well annotated datasets, such as COCO \cite{lin2014microsoft} and Pascal VOC \cite{Everingham10}. However, the difficulties in building larger detection datasets hinder the further development of SOD.  Weakly supervised object discovery (WSOD) \cite{wei2019unsupervised,zhang2020rethinking,zhou2016learning,selvaraju2017grad} is able to coarsely localize objects of interest with only image-level labels, which makes WSOD applicable to more large-scale datasets without bounding box annotations, such as ImageNet \cite{deng2009imagenet}. Class activation map (CAM) \cite{zhou2016learning} is able to visualize the importance of different regions of the input image by utilizing the output of the pooling layer at the end of CNNs. Grad-CAM \cite{selvaraju2017grad} further generalizes CAM by using the gradients to calculate the importance of different pixels, which makes it applicable to more CNN architectures. ACoL \cite{zhang2018adversarial} designs a CNN architecture with two branches to adversarially learn the full region of objects. CutMix \cite{yun2019cutmix} randomly blends different regions of images to improve the localization accuracy. However, the huge computational cost and latency make both the SOD (e.g., SSD) and WSOD inapplicable in our context, i.e., resource-constrained embedded hardware.

\noindent
\textbf{CNN compression: }
To compress CNNs for better on-device performance, different approaches have been proposed to remove the redundancy of CNNs. Deep Compression \cite{han2015deep} removes neurons from kernels, which achieves a high compression rate. However, the obtained sparse networks need special hardware to accelerate. \cite{lin2020hrank,molchanov2019importance,sandler2018mobilenetv2,yu2019slimmable} build compact CNNs by removing channels from each layer, which can speed up the execution of CNNs on off-the-shelf hardware. Among them MobileNetV2 \cite{sandler2018mobilenetv2} and Slimmable networks \cite{yu2019slimmable} uniformly reduce the channels in all layers. Instead, HRank \cite{lin2020hrank} and Taylor pruning \cite{molchanov2019importance} prune the channels in a layer-wise manner. Besides compressing the network architecture, some works focus on reducing the spatial redundancy in images. MobileNetV2 \cite{sandler2018mobilenetv2} and MnasNet \cite{tan2019mnasnet} use smaller resolutions to improve the performance on mobile devices. DR-ResNet \cite{zhu2021dynamic} proposes a dynamic resolution network to reduce the computational cost. GFNet \cite{wang2020glance} employs reinforcement learning to select multiple small patches instead of using the whole image to accelerate CNNs. However, these methods only consider reducing the redundancy in a single dimension and thus only achieve a limited compression rate. In contrast, jointly compressing all dimensions promises a better trade-off between the compression rate and accuracy.

\begin{figure*}
    \centering
    \includegraphics[width=0.95\textwidth]{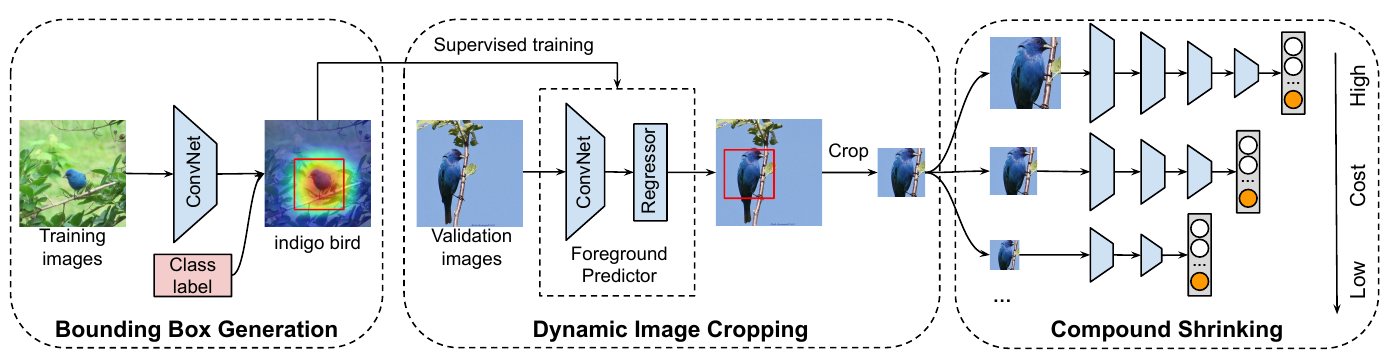}
    \caption{The overview of the proposed framework, Smart Scissor (SS), which mainly consists of three components: bounding box generation (BBG), dynamic image cropping (DIC), and compound shrinking (CS).}
    \label{fig:framework}
\end{figure*}

\section{The Smart Scissor Framework}
\label{sec:method}

In this section, we first outline the design of \ssr and then describe each submodule in detail.

As demonstrated in Figure \ref{fig:framework}, given a classification dataset $\mathcal{D}$ and a classification network $\mathcal{N}$. We first exploit Grad-CAM to generate the salience map of all training images in $\mathcal{D}$, and then we generate a bounding box for each image according to the salience map and form a bounding box label set $\mathcal{B}$. Thereafter, we utilize the image-box pairs \{$\mathcal{D}_i$, $\mathcal{B}_i$\} to train a lightweight predictor. Meanwhile, for a given MACs budget $\mathcal{M}$, we use CS to quickly calculate the desired resolution and shrink the network accordingly. During inference, an image will be first fed into the predictor to quickly output the boundary of the foreground object, then the detected object is cropped. Finally, the cropped patch is resized to the calculated resolution and sent to the compressed network $\mathcal{N}_s$ for fast inference. 

\subsection{Bounding Box Generation}
\label{subsec:generate}

% As aforementioned, For quite a few images, the redundant background occupies much computation and the foreground object that really contributes to the final prediction only takes a few pixels. Inspired by object detection tasks \cite{he2017mask,ren2015faster,liu2016ssd}, we propose to localize the foreground object and only conduct computation on the localized area, which reduces redundant computation on the background while preserving the important features of the foreground object.

As aforementioned, dynamically cropping the foreground for inference is promising in reducing computation and improving accuracy. However, for classification datasets like ImageNet, there is no out-of-the-box position annotation for the foreground object. Moreover, the position of the foreground object varies in different images, which makes it difficult to efficiently localize the foreground object.

% Traditional supervised object detection (SOD) methods are not applicable for our task, because the largest SOD dataset, COCO \cite{lin2014microsoft}, only contains 80 classes and is unable to generalize to ImageNet-1K which has 1000 classes. 

To address this limitation, we first use Grad-CAM \cite{selvaraju2017grad} to automatically generate the position annotations. 
% As shown in Figure \ref{fig:framework}, given an image and the corresponding class label, Grad-CAM is capable of visualizing the importance of different regions to the final prediction. 
Specifically, let the class label of the given image be $c$. We first perform forward inference with a well-trained CNN (e.g. ResNet50) to obtain the prediction score $p^c$ for class $c$, and then conduct backpropagation to compute the gradient of the score $p^c$ with respect to each activation of the last convolutional layer. Thereafter, the gradients are aggregated within each channel via global average pooling. The obtained scalar for each channel can be seen as the weight of the channel, which can be calculated as follows:
\begin{equation}
    a_k^c = \overbrace{\frac{1}{Z}\sum_i\sum_j}^{\textit{pooling}} \overbrace{\frac{\partial p^c}{A_{ij}^k}}^{\textit{gradients}}
\end{equation}
where $a_k^c$ is the weight of channel $k$ for class $c$, and $A_{ij}^k$ is a single activation indexed by $i$ and $j$ in the 2-D feature map of channel $k$. With the weights of all channels determined, the salience map for class $c$ can be obtained by computing the weighted sum of all feature maps over the channel dimension, which is formulated as:
\begin{equation}
    L_{Grad\_CAM}^c = ReLU\underbrace{(\sum_k a_k^c A^k)}_{\textit{linear combination}}
\end{equation}
where $A^k$ is the 2-D feature map of channel $k$, and ReLU is used to eliminate the impact of negative activations. Finally, the obtained salience map is upsampled to the same size as the input image via bi-linear interpolation algorithm.

\begin{figure}
    \centering
    \includegraphics[width=0.47\textwidth]{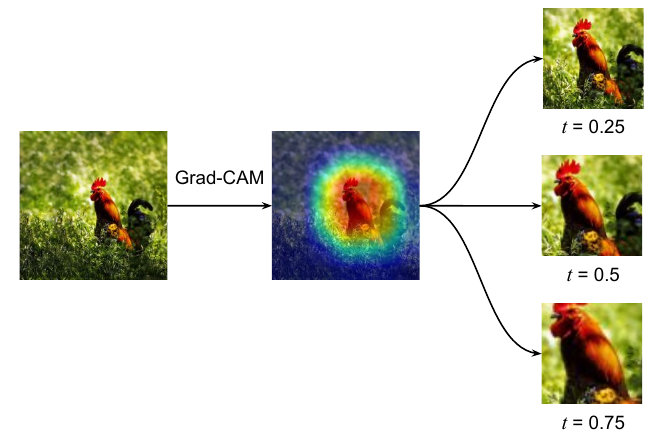}
    \caption{By applying different values of the salience threshold $t$, we can obtain different cropped images. The larger the threshold value, the more radical the cropping.}
    \label{fig:threshold}
\end{figure}

With the salience map generated, we then introduce a simple yet effective strategy to determine the bounding box of the foreground object. Initially, we set the box as the boundary of the image. Subsequently, we shrink the four sides of the box simultaneously, and once a side reaches our preset salience threshold $t$, the side is frozen. The bounding box is determined after all sides are frozen. Note that it is crucial to appropriately select the value of $t$ for the final result. As demonstrated in Figure \ref{fig:threshold}, a too small threshold will result in residual background redundancy, while a too large threshold will lose some object features. Therefore, we conduct empirical experiments to determine the optimal threshold value. As shown in Table \ref{tab:threshold}, we achieve the highest accuracy when the threshold $t$ is set to 0.5. Therefore, we set $t=0.5$ in the following experiments. Note that more fine-grained searching for $t$ may further improve the accuracy, but it also increases the search cost. Finally, the generated box annotations are saved in the form of [$X_{min}$, $Y_{min}$, $X_{max}$, $Y_{max}$], which denotes the position of the foreground in the image. 

\begin{table}[tbp]
  \centering
  \caption{The impact of using different salience thresholds on prediction accuracy. The model is trained and evaluated on ImageNet-1K. $t=0$ means using the original images without Grad-CAM cropping.}
    \begin{tabular}{l||c||c||c||r}
    \toprule
    \textbf{Model} & \textbf{$\#$Params} & \textbf{$\#$MACs} & \textbf{\textit{t}} & \textbf{Acc@1 } \\
    \midrule
    \multirow{4}[2]{*}{ResNet50} & \multirow{4}[2]{*}{25.6 M} & \multirow{4}[2]{*}{4.1 B} & 0 (Baseline) & 76.02 $\%$ \\
        & &  & 0.25  & 76.45 $\%$ \\
        & &  & 0.5   & \textbf{76.88 $\%$} \\
        & &  & 0.75  & 76.32 $\%$ \\
    \bottomrule
    \end{tabular}%
  \label{tab:threshold}%
\end{table}%

\subsection{Dynamic Image Cropping}
\label{subsec:predict}

\begin{figure}[tbp]
    \centering
    \includegraphics[width=0.47\textwidth]{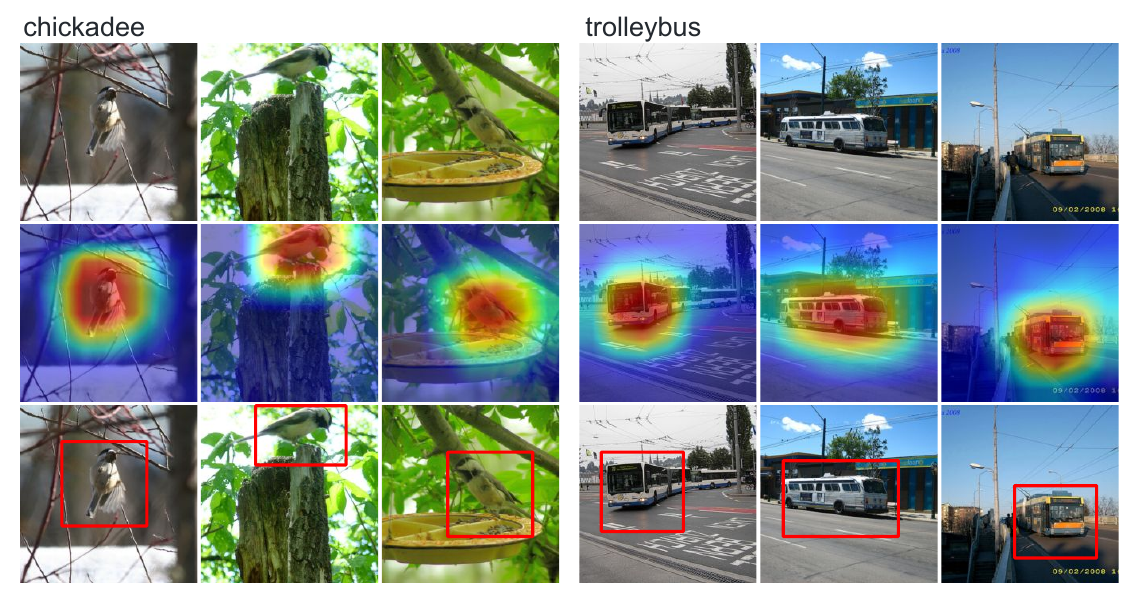}
    \caption{The bounding boxes generated with the salience threshold $t=0.5$, which accurately localize the key object in each image.}
    \label{fig:grad-cam}
\end{figure}

Figure \ref{fig:grad-cam} shows that we are capable of accurately localizing the foreground of images with Grad-CAM. However, Grad-CAM cannot be directly applied to edge applications because of the time-consuming backpropagation process. Moreover, Grad-CAM requires the class label as weak supervision, which is unavailable for validation images. To address these issues, we design a foreground predictor to efficiently localize the foreground of input images.

\subsubsection{Predictor Architecture}

\begin{figure}[htbp]
    \centering
    \includegraphics[width=0.47\textwidth]{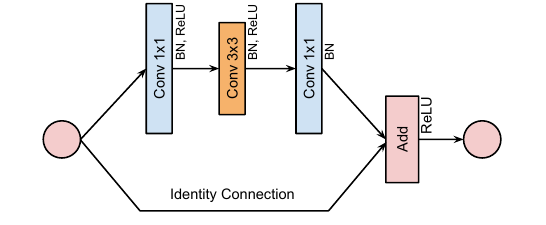}
    \caption{The architecture of the residual bottleneck block, which is widely used in many state-of-the-art networks.}
    \label{fig:block}
\end{figure}

Although dozens of object detectors have been proposed to detect objects from images, such as SSD \cite{liu2016ssd} and Faster R-CNN \cite{ren2015faster}, these detector architectures are completely inapplicable to our task. Particularly, to accurately detect multiple objects from a single image, existing detectors are usually equipped with a heavy feature extractor and a complex region proposal network (RPN), which are extremely time-consuming and unnecessary for image classification where each image only contains one object. Therefore, we discard the design of traditional detectors. Instead, we design the foreground predictor as a lightweight plain CNN, whose detailed architecture is summarized in Table \ref{tab:arch}. It consists of several residual bottleneck blocks \cite{he2016deep} and a fully connected layer as the single-box regressor. As shown in Figure \ref{fig:block}, a residual bottleneck contains two convolutional layers with 1$\times$1 kernels and one convolutional layer with 3$\times$3 kernels in the middle. The computational cost mainly results from the 3$\times$3 convolutional layer. Therefore, to reduce the cost and accelerate the predictor, we only stack two residual bottleneck blocks in each stage and each block is only equipped with a small number of channels. Consequently, the proposed predictor only contains 0.27M parameters and 0.09B MACs, which is negligible compared to popular object detectors (e.g., Faster R-CNN with 134.7M (499$\times$) parameters and 15.1B (167.8$\times$) MACs \cite{li2018tiny}). Moreover, the small overhead of the foreground predictor can be complemented by the CS module.

\subsubsection{Training of Foreground Predictor}
We train the predictor in a supervised manner. First, we generate a bounding box label set $\mathcal{B}$ for all training images as described in Subsection \ref{subsec:generate}, then the labels are utilized to train the predictor. We use the mean square error (MSE) as the loss function. Let $\mathcal{P}_i=$ [$X_{min}^p$, $Y_{min}^p$, $X_{max}^p$, $Y_{max}^p$] be the output of the predictor, and $\mathcal{B}_i=$ [$X_{min}^g$, $Y_{min}^g$, $X_{max}^g$, $Y_{max}^g$] be the generated box label, the loss function can be formulated as:
\begin{equation}
\begin{aligned}
    \mathcal{L}_{box}  = & \ MSELoss(\mathcal{P}_i, \ \mathcal{B}_i) \\
                       = & \ \frac{1}{4} ((X_{min}^g-X_{min}^p)^2+(Y_{min}^g-Y_{min}^p)^2 \\
                                & +(X_{max}^g-X_{max}^p)^2+(Y_{max}^g-Y_{max}^p)^2)
\end{aligned}
\end{equation}

To balance the training overhead and prediction accuracy, we train the predictor with Adam \cite{kingma2015adam} optimizer for 40 epochs. The initial learning rate is set to 1e-3, and the learning rate is scheduled using exponential decay \cite{Li2020An}. The training of the box predictor is decoupled with backbone networks. Once the predictor is trained, it can be directly applied to different classification backbones without any training overhead. During inference, the trained predictor will quickly localize the foreground object of the input image and generate a finely cropped image, which significantly reduces the redundancy in the input image.

\subsection{Compound Shrinking}
\label{subsec:compound}

The proposed DIC significantly reduces the redundancy in images, improving the computational efficiency. We observe that redundancy also exists in network architectures (e.g., redundant parameters), and only removing the redundancy in images loses the opportunity to further compress the model for embedded hardware. Besides, \cite{tan2019efficientnet} demonstrates that jointly adjusting different dimensions promises higher accuracy. To this end, we propose a compound shrinking (CS) strategy to jointly compress the three dimensions (depth, width, resolution) of CNNs to reduce the redundancy in images as well as networks while maintaining the accuracy.

\begin{table}[htbp]
  \centering
  \caption{The architecture of the proposed box predictor. \#C denotes the number of channels and \#L denotes the number of layers.}
    \begin{tabular}{lllll}
    \toprule
    \multicolumn{1}{c||}{\textbf{Stage}} & \multicolumn{1}{c||}{\textbf{Block}} & \multicolumn{1}{c||}{\textbf{Resolution}} & \multicolumn{1}{c||}{\textbf{\#C}} & \multicolumn{1}{c}{\textbf{\#L}} \\
    \midrule
    \multicolumn{1}{c||}{1} & \multicolumn{1}{c||}{Conv 3$\times$3} & \multicolumn{1}{c||}{224 $\times$ 224} & \multicolumn{1}{c||}{16} & \multicolumn{1}{c}{1} \\
    \multicolumn{1}{c||}{2} & \multicolumn{1}{c||}{Residual Bottleneck} & \multicolumn{1}{c||}{112 $\times$ 112} & \multicolumn{1}{c||}{16} & \multicolumn{1}{c}{2} \\
    \multicolumn{1}{c||}{3} & \multicolumn{1}{c||}{Residual Bottleneck} & \multicolumn{1}{c||}{56 $\times$ 56} & \multicolumn{1}{c||}{32} & \multicolumn{1}{c}{2} \\
    \multicolumn{1}{c||}{4} & \multicolumn{1}{c||}{Residual Bottleneck} & \multicolumn{1}{c||}{28 $\times$ 28} & \multicolumn{1}{c||}{32} & \multicolumn{1}{c}{2} \\
    \multicolumn{1}{c||}{5} & \multicolumn{1}{c||}{Residual Bottleneck} & \multicolumn{1}{c||}{14 $\times$ 14} & \multicolumn{1}{c||}{64} & \multicolumn{1}{c}{2} \\
    \multicolumn{1}{c||}{6} & \multicolumn{1}{c||}{Pooling \& Linear} & \multicolumn{1}{c||}{7 $\times$ 7} & \multicolumn{1}{c||}{4} & \multicolumn{1}{c}{1} \\
    \midrule
    \multicolumn{5}{l}{\#Params: 0.27M} \\
    \multicolumn{5}{l}{\#MACs: 0.09B} \\
    % \multicolumn{5}{l}{Latency: 9.2ms} \\
    \bottomrule
    \end{tabular}%
  \label{tab:arch}%
\end{table}%

Intuitively, shrinking different dimensions has different impacts on accuracy and model overhead. The core of our compound shrinking strategy is to calculate a shrinking coefficient for each dimension according to their trade-off between accuracy and model overhead. A larger coefficient denotes more radical shrinking. More specifically, the dimension with a steep accuracy drop during shrinking will be assigned a small shrinking coefficient to prevent severe accuracy degradation. To calculate the shrinking coefficients, we first quantify the trade-off of each dimension between accuracy and model overhead. Here we use MACs as the metric to measure the cost of models, because all three dimensions are related to the MACs of a model while only the depth and width can affect the model parameters. Given a MACs budget $\mathcal{M}$, we first obtain the accuracy drops resulting from separately shrinking different dimensions, which can be represented as:
\begin{equation}
\label{eq:acc}
    \Delta A_s(\mathcal{M}) = A_0 - A_s(\mathcal{M})
\end{equation}
where $s\in\{d, w, r\}$ represents the shrunk dimension, $A_s(\mathcal{M})$ denotes the accuracy of the shrunk model, and $A_0$ denotes the accuracy of the original model. To comply with the rule that the steeper the drop in accuracy, the smaller the coefficient of the corresponding dimension, we design the following equation to determine the shrinking coefficient for each dimension:
\begin{equation}
\label{eq:coeff}
    \mathcal{C}_s(\mathcal{M}) = \frac{\sqrt[3]{\Delta A_d(\mathcal{M}) \cdot \Delta A_w(\mathcal{M}) \cdot \Delta A_r(\mathcal{M})}}{\Delta A_s(\mathcal{M})}
\end{equation}
where $\mathcal{C}_s(\mathcal{M})$ denotes the shrinking coefficient of the dimension $s$ ($s\in\{d, w, r\}$). Through Equation \ref{eq:acc} and Equation \ref{eq:coeff}, we are able to efficiently calculate the coefficients once we obtain the accuracy degradation of the three dimensions in the given MACs regime. 

\begin{figure}
    \centering
    \includegraphics[width=0.48\textwidth]{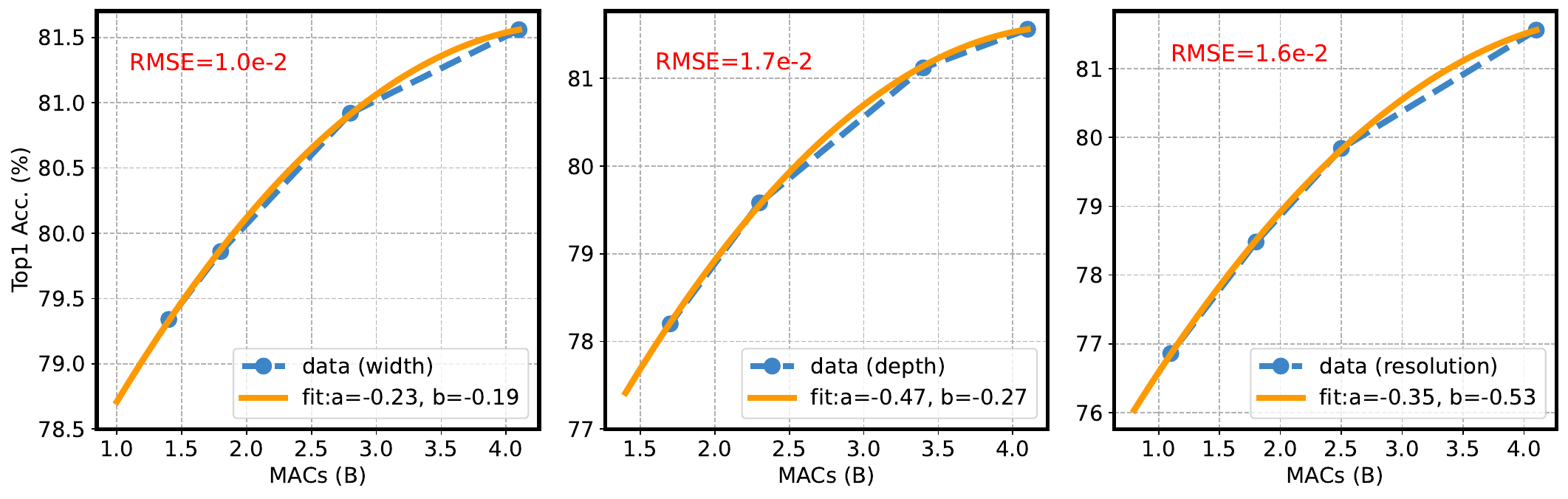}
    \caption{The actual accuracy (blue dotted line) and the estimated accuracy (yellow line) over MACs by separately shrinking the three dimensions. The low root mean square error (RMSE) indicates that the accuracy estimator can well fit existing data.}
    \label{fig:dimensions}
\end{figure}

However, the training cost of the compressed models to calculate the accuracy drop is still non-negligible. To mitigate the training overhead, we propose a dimension-wise accuracy estimator to quickly estimate the accuracy of the compressed models and calculate the accuracy degradation resulting from shrinking different dimensions in the given MACs regime. First, we sample a couple of models with different MACs by separately shrinking the three dimensions. As demonstrated in Figure \ref{fig:dimensions}, the accuracy distribution of the three dimensions along MACs can be well fitted by a quadratic polynomial. Therefore, we design a simple yet effective polynomial estimator to predict the accuracy with respect to the target MACs $\mathcal{M}$. The estimator can be formulated as follows:
\begin{equation}
    A_s(\mathcal{M}) = a_s (\mathcal{M} - \mathcal{M}_0)^2 + b_s (\mathcal{M} - \mathcal{M}_0) + A_0 
\end{equation}
where $\mathcal{M}_0$ is the MACs of the original model. $a_s$ and $b_s$ are the hyperparameters to fit for dimension $s$ ($s\in\{d, w, r\}$). Subsequently, we train the dimension-wise estimator using least square regression with the aforementioned sampled data. Figure \ref{fig:dimensions} shows that the proposed estimator can well fit existing data. Due to the simple and intuitive design of the estimator, we only need to sample and train very few models to train the estimator, and this cost is a one-time cost. With the accuracy estimator established, we are capable of directly calculating the accuracy drop and subsequently the shrinking coefficient of each dimension across a wide range of MACs regimes. Consequently, the cost of assigning the coefficients is significantly reduced compared to directly training models to obtain the coefficients.

% To find out the trade-off of each dimension, we first separately shrink each dimension to different MACs regimes and fully train the shrunk models to obtain the accuracy. We visualize the trade-off of each dimension in Figure \ref{fig:dimensions}, which reveals that the accuracy declines parabolically along model MACs for all dimensions.

\section{Experimental Results}
\label{sec:exp}

In this section, we conduct experiments to comprehensively evaluate \ssr. We compare \ssr with many state-of-the-art (\sota) methods in terms of accuracy, MACs, on-device latency, and throughput. In addition, we also do ablation experiments to validate the efficacy of each component in our framework.

\subsection{Hardware Devices}
\label{subsec:hardware}
The on-device latency and throughput are two important metrics to evaluate the efficiency of CNNs. To demonstrate the on-device performance of our method, we select two representative embedded GPU platforms, NVIDIA AGX Xavier and NVIDIA Jeton Nano, and Intel i7-9750H@2.6GHz CPU to deploy the proposed framework and measure the latency and throughput. 
% NVIDIA AGX Xavier is an integrated powerful edge GPU platform, which includes a Volta GPU with 512 CUDA cores and can provide a computing power of up to 11 TOPS (INT8) with a power consumption of 15W. NVIDIA Jetson Nano is an energy-efficient edge GPU platform, which consumes much less energy than Xavier while still providing considerable performance. Intel i7-9750H is a desktop CPU which is widely used in personal computers.
The hardware specifications are listed in Table \ref{tab:device}.

\begin{table}[tbp]
  \centering
  \caption{The specifications of three different hardware platforms used in our experiments.}
    \begin{tabular}{l||c||c||r}
    \toprule
    \textbf{Device} & \textbf{Power} & \textbf{Memory/Cache} & \textbf{Performance} \\
    \midrule
    % \multirow{3}[1]{*}{Xavier} & 10 W  & \multirow{3}[1]{*}{32 GB} & 8 TOPS \\
    %       & 15 W  &       & 11 TOPS \\
    %       & 30 W  &       & 22 TOPS \\
    AGX Xavier & 15 W & 32 GB & 11 TOPS \\
    \midrule
    Jetson Nano  & 5 W   & 4 GB  & 0.47 TFLOPS \\
    \midrule
    i7-9750H & 45 W  & 12 MB & 0.4 TOPS \\
    \bottomrule
    \end{tabular}%
  \label{tab:device}%
\end{table}%

\subsection{Datasets}
\label{subsec:dataset}
We evaluate our method on two large-scale datasets, ImageNet-1K \cite{deng2009imagenet} and ImageNet-100. ImageNet-1K (a.k.a. ILSVRC 2012) is the most widely exploited dataset for image classification tasks, which consists of 1,000 classes, and each class contains about 1,000 high-resolution (224$\times$224) images for training and 50 images for validation. ImageNet-100 is a subset of ImageNet-1K, which is composed by 100 randomly selected classes from ImageNet-1K. The detailed categories of ImageNet-100 is provided in the code repository. All images are preprocessed following a simple configuration as \cite{radosavovic2020designing,sandler2018mobilenetv2}.

\subsection{Networks}
\label{subsec:network}
We implement the \ssr framework with two popular CNN models, ResNet50 \cite{he2016deep} and RegNet-X \cite{radosavovic2020designing} as the baseline backbone networks. For each model, we apply \ssr to reduce the model overhead by removing both the spatial redundancy of images and the architecture redundancy of networks. Moreover, we also report the results of other \sota model compression approaches on the two models for comparison.

\subsection{Optimization Settings}
\label{subsec:setting}

We use a stochastic gradient descent (SGD) optimizer with a momentum of 0.9 to train the classification networks. On both ImageNet-100 and Imagenet-1K, models are first trained with a batch size of 1,024 for 100 epochs without dynamic image cropping, where the first 5 epochs are for warmup. At this stage, the initial learning rate is set to 2.0 and decayed by exponential learning rate policy with a decay factor of 0.02. Then, we use the proposed dynamic image cropping to fine-tune the pretrained models for 20 epochs with a constant learning rate of 5e-4. To prevent over-fitting, we also use label smoothing with $\epsilon=0.1$. 
% For large models, we reduce the batch size and the learning rate proportionally as in \cite{goyal2017accurate} to save the usage of CUDA memory.

\subsection{Results on ImageNet-100}
We conduct experiments on ImageNet-100 with both ResNet50 and RegNet-X. In practise, to better demonstrate the improvement of each component, we implement two versions of Smart Scissor, Smart Scissor with only the DIC module (\ssd) and Smart Scissor with both the DIC and the CS module (\css). As a comparison, we use the most popular image cropping method, ResizedCenterCrop (\rcc) to crop and resize images. Finally, all models are deployed to the hardware platforms to evaluate the latency and throughput.

\begin{figure}
    \centering
    \includegraphics[width=0.48\textwidth]{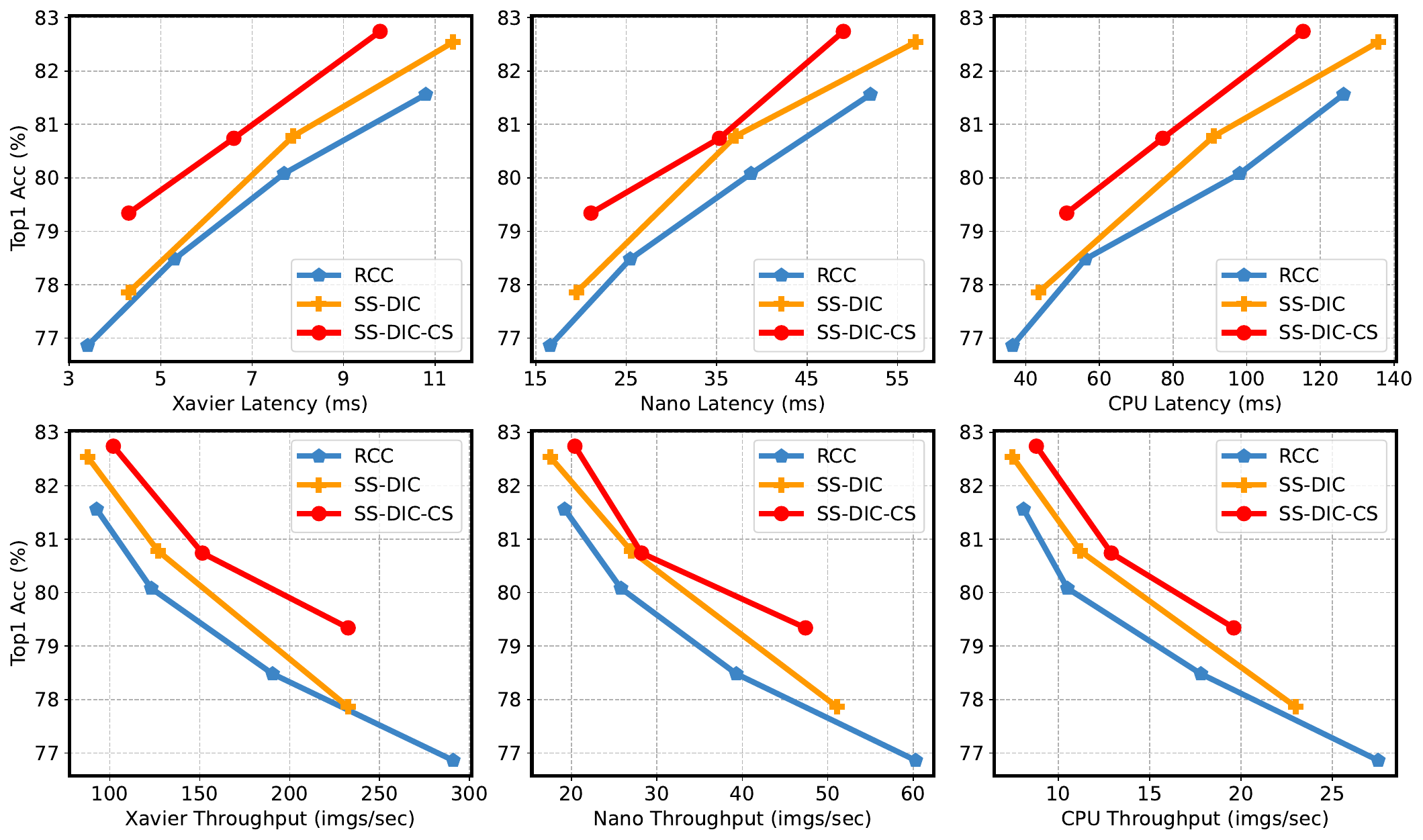}
    \caption{The real performance of ResNet50 compressed by different methods on three distinct hardware devices. The accuracy is measured on ImageNet-100.}
    \label{fig:res50_perf}
\end{figure}

\begin{table}[tbp]
  \centering
  \caption{Results of ResNet50 on ImageNet-100.}
    \begin{tabular}{l||r||r||r||r}
    \toprule
    \textbf{Method} & \textbf{\#Params} & \textbf{\#MACs} & \textbf{$\downarrow$ MACs}& \textbf{ Acc@1} \\
    \midrule
    \rcc-Baseline & 23.7 M  & 4.1 B & 0.0$\%$  & 81.6$\%$  \\
    \ssd    & 24.0 M  & 4.2 B & -2.4$\%$  & 82.5$\%$  \\
    \textbf{\css} & \textbf{17.3 M } & \textbf{3.0 B } & \textbf{26.8$\%$} & \textbf{82.7$\%$ } \\
    \midrule
    \rcc   & 23.7 M  & 3.0 B  & 26.8$\%$ & 80.1$\%$  \\
    \ssd    & 24.0 M  & 2.6 B  & 36.6$\%$  & 80.8$\%$  \\
    \textbf{\css} & \textbf{14.5 M } & \textbf{2.4 B } & \textbf{41.5$\%$} & \textbf{81.5$\%$ } \\
    \midrule
    \rcc   & 23.7 M  & 1.1 B & 73.2$\%$   & 76.9$\%$  \\
    \ssd   & 24.0 M  & 1.2 B & 70.7$\%$  & 77.9$\%$  \\
    \textbf{\css} & \textbf{7.8 M } & \textbf{1.0 B } & \textbf{75.6$\%$} & \textbf{79.3$\%$ } \\
    \bottomrule
    \end{tabular}%
  \label{tab:res-100}%
\end{table}%

% \noindent
\textbf{ResNet50: }As shown in Table \ref{tab:res-100}, \css outperforms the competitors on all metrics. Specifically, compared to the baseline ResNet50 (\rcc-Baseline), \ssd improves the accuracy by 0.9\% with a negligible increase in model parameters (1.2\%) and MACs (2.4\%), while \css further pushes up the accuracy improvement to 1.1\% with a parameter reduction of 27\% and a MACs reduction of 26.8\%. In the low 
complexity regime, \css achieves 2.4\% higher accuracy than \rcc with only 33\% model parameters (7.8 M v.s. 23.7 M) and less MACs. Meanwhile, Figure \ref{fig:res50_perf} indicates that both \ssd and \css outperform \rcc by a large margin across a wide spectrum in terms of latency and throughput. Particularly, \css achieves 79.4\% top-1 accuracy with a latency of 4.3 ms on Xavier, which is 0.9\% higher in accuracy and 18.9\% lower in latency compared to \rcc (78.5\% top1 accuracy, 5.3 ms). At the same time, the throughput of \css on Xavier is 232.5 \textit{imgs/sec}, which is 22\% higher than \rcc (190.5 \textit{imgs/sec}). On Nano and Intel i7-9750H CPU, \css also improves the throughput by 20.6\% and 10.1\%, and reduces the latency by 17\% and 9.1\%, respectively.

\begin{figure}
    \centering
    \includegraphics[width=0.48\textwidth]{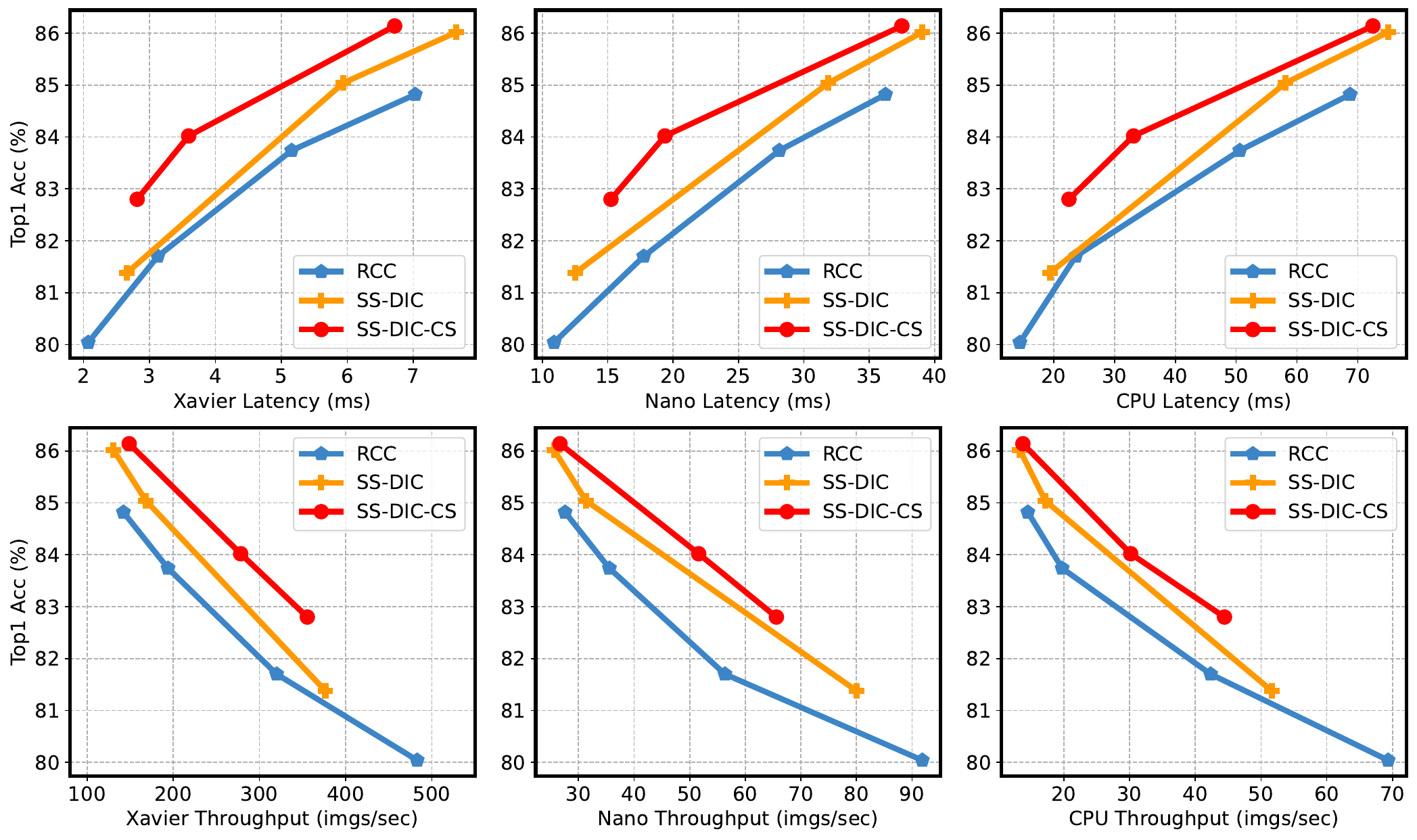}
    \caption{The real performance of RegNet-X compressed by different methods on three distinct hardware devices. The accuracy is measured on ImageNet-100.}
    \label{fig:reg_perf}
\end{figure}

\begin{table}[htbp]
  \centering
  \caption{Results of RegNet-X on ImageNet-100.}
    \begin{tabular}{l||r||r||r||r}
    \toprule
    \textbf{Method} & \textbf{\#Params} & \textbf{\#MACs} & \textbf{$\downarrow$ MACs} & \textbf{Acc@1} \\
    \midrule
    \rcc-Baseline & 8.4 M & 1.6 B & 0.0\% & 84.8\% \\
    \ssd & 8.7 M & 1.3 B & 18.8\% & 85.0\% \\
    \textbf{\css} & \textbf{6.3 M} & \textbf{1.3 B} & \textbf{18.8\%} & \textbf{86.1\%} \\
    \midrule
    \rcc   & 8.4 M & 0.7 B & 56.3\% & 82.3\% \\
    \ssd & 8.7 M & 0.8 B & 50.0\% & 83.8\% \\
    \textbf{\css} & \textbf{3.6 M} & \textbf{0.6 B} & \textbf{62.5\%} & \textbf{84.0\%} \\
    \midrule
    \rcc   & 8.4 M & 0.4 B & 75.0\% & 80.0\% \\
    \ssd & 8.7 M & 0.5 B & 68.8\% & 81.4\% \\
    \textbf{\css} & \textbf{2.5 M} & \textbf{0.4 B} & \textbf{75.0\%} & \textbf{82.8\%} \\
    \bottomrule
    \end{tabular}%
  \label{tab:regnet}%
\end{table}%

\textbf{RegNet-X: } The experimental results of RegNet-X are summarized in Table \ref{tab:regnet}, where we also observe a significant improvement of our method. \css outperforms the baseline RegNet-X (RCC-Baseline) with an improvement of 1.3\% in accuracy and a reduction of 18.8\% in MACs. Meanwhile, \css reduces the model parameters by 25\% (6.3 M v.s. 8.4 M). In the low MACs regime, \css observes a remarkable 2.8\% improvement in accuracy with only 30\% parameters (2.5 M v.s. 8.4 M) compared to \rcc. As for the real performance on hardware, \css obtains an accuracy of 84\% with 3.6 ms latency on Xavier, which is 0.3\% higher in accuracy and 30.8\% lower in latency than \rcc (83.7\% top-1 accuracy, 5.2 ms). Similarly, the latency reductions on Nano and Intel CPU are 31\% and 34.6\%, respectively. Besides, \css observes a 44.9\% throughput improvement (51.6 \textit{imgs/sec} v.s. 35.6 \textit{imgs/sec}) on Nano and a 52.5\% throughput improvement (30.2 \textit{imgs/sec} v.s. 19.8 \textit{imgs/sec}) on Intel CPU compared to \rcc.

\begin{figure}
    \centering
    \includegraphics[width=0.4\textwidth]{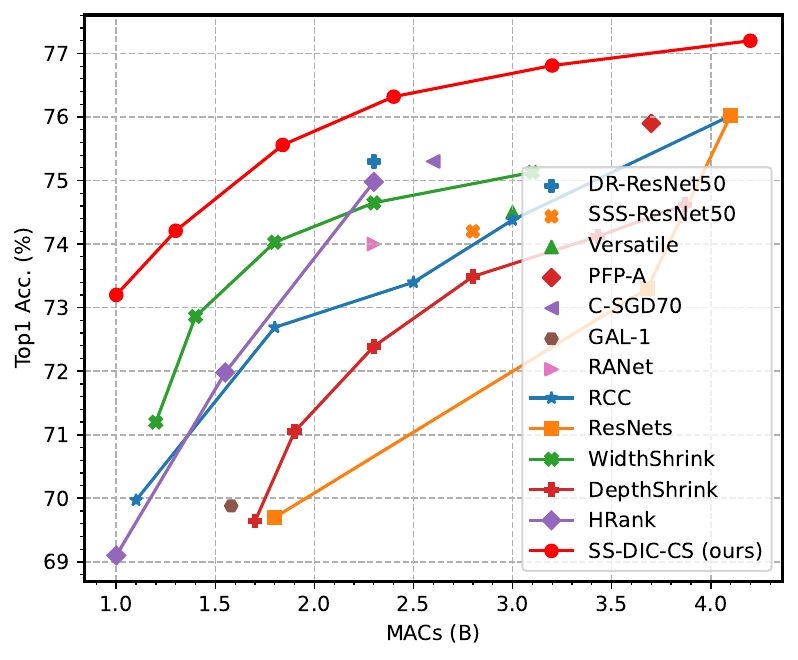}
    \caption{Comparison of our Smart Scissor with other state-of-the-art model compression methods. The baseline model is ResNet50 and the dataset is ImageNet-1K.}
    \label{fig:res-1k}
\end{figure}

\subsection{Results on ImageNet-1K}
We evaluate our approach on ImageNet-1K, and compare the evaluation results with many SOTA CNN compression frameworks. In addition, we also compare our compressed models with many popular backbone architectures in different computation regimes.

\textbf{Comparison with \sota compression methods: }Starting with the baseline ResNet50 model, we implement different model shrinking methods, including \rcc, width shrinking (WidthShrink) \cite{sandler2018mobilenetv2, zagoruyko2016wide}, depth shrinking (DepthShrink) \cite{he2016deep}, \ssd, and \css, to compress the three dimensions of the model to different MACs regimes and compare their performance. In addition, we also report the performance of multiple \sota model compression techniques from the related papers, including DR-ResNet50 \cite{zhu2021dynamic}, SSS-ResNet50 \cite{huang2018data}, Versatile-ResNet50 \cite{wang2018learning}, PFP-A-ResNet50 \cite{liebenwein2019provable}, C-SGD70-ResNet50 \cite{ding2019centripetal}, GAL-1 \cite{lin2019towards}, HRank \cite{lin2020hrank}, and RANet \cite{yang2020resolution}. The comparison results are summarized in Figure \ref{fig:res-1k}, which shows that our method achieves the highest accuracy across a wide range of MACs. Particularly, compared to the baseline ResNet50, our method achieves 1.2\% accuracy improvement (76.0\% to 77.2\%) with a negligible increase in MACs (4.1 B to 4.2 B). As we continue to reduce the MACs budget, \css reduces the MACs by 41.5\% (4.1 B to 2.4 B) while still achieving 0.3\% higher accuracy (76.0\% to 76.3\%). In the low MACs regime, \css remarkably improves the accuracy by 4.2\% (70.0\% to 74.2\%) compared to \rcc with similar MACs. In comparison with other \sota compression methods, our method also achieves the best trade-off between MACs and accuracy. For example, \css achieves 3.3\% higher accuracy (73.2\% v.s. 69.9\%) than GAL-1 \cite{lin2019towards}  with 37.5\% less MACs (1.0B v.s. 1.6B).

\begin{figure*}
    \centering
    \includegraphics[width=0.97\textwidth]{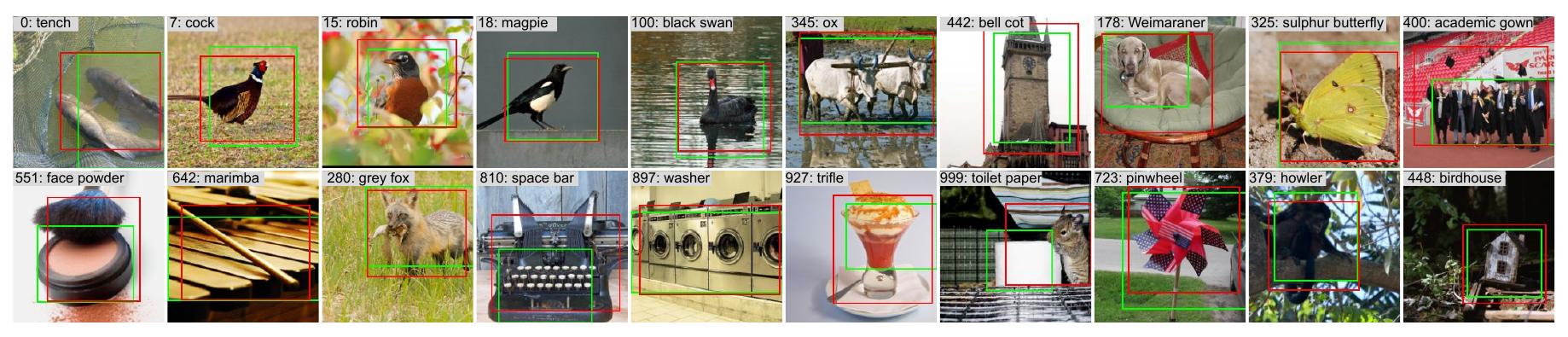}
    \caption{Visualization of the predicted bounding boxes (red) and the ground truth bounding boxes generated from Grad-CAM (green). Our predictor achieves a high localization accuracy of 62\% mAP on ImageNet-1K validation set. The images above are randomly selected from ImageNet-1K.}
    \label{fig:visualbox}
\end{figure*}

\textbf{Comparison with popular backbones: }
In this experiment, we apply \ssr to ResNet50 and compare it with other models from the ResNet family, such as ResNet101 and ResNet34, etc. In addition, we also conduct extensive comparison with other popular backbones like DenseNets \cite{huang2017densely} and the Inception family \cite{szegedy2016rethinking, ioffe2015batch}. As demonstrated in Table \ref{tab:backbones}, in the highest MACs regime, \ssd only uses 53.2\% of the MACs (4.2 B v.s. 7.9 B) of DenseNet161 to achieve the same level of accuracy, while in the lowest MACs regime, \css obtains the highest top-1 accuracy (75.6\%), which is 5.8\% and 2.1\% higher than ResNet18 (69.8\%) and BN-Inception (73.5\%), respectively. The comparison results with other backbones reveal that our method can achieve promising results without redesigning the network architecture, which avoids the extremely time-consuming exploration of the architecture design space.

\begin{table}[tbp]
  \centering
  \caption{Comparison with other popular backbone networks.}
    \resizebox{0.47\textwidth}{!}{
        \begin{tabular}{l||r||r||r||r}
        \toprule
        \textbf{Model} & \textbf{\#Params} & \textbf{\#MACs} & \textbf{$\downarrow$ MACs} & \textbf{Acc@1} \\
        \midrule
        ResNet50\cite{he2016deep} & 25.6 M & 4.1 B &   0.0$\%$   & 76.0\% \\
        \midrule
        ResNet101\cite{he2016deep} & 44.6 M & 7.9 B &    -92.7$\%$   & \textbf{77.4\%} \\
        DenseNet161\cite{huang2017densely} & 28.7 M & 7.9 B &   -92.7$\%$    & 77.1\% \\
        InceptionV3\cite{szegedy2016rethinking} & 27.2 M & 5.8 B &   -41.5$\%$    & 77.3\% \\
        \textbf{\ssd} & \textbf{25.9 M} & \textbf{4.2 B} &   \textbf{-2.4$\%$}    & 77.2\% \\
        \midrule
        ResNet34\cite{he2016deep} & 21.8 M & 3.7 B &    9.8$\%$   & 73.3\% \\
        DenseNet169\cite{huang2017densely} & \textbf{14.2 M} & 3.4 B &   17.1$\%$    & 75.6\% \\
        \ssd & 25.9 M & 3.1 B &   24.4$\%$      & 76.3\% \\
        \textbf{\css} &  15.4 M     &  \textbf{2.4 B}     &    \textbf{41.5$\%$}   & \textbf{76.3 $\%$}  \\
        \midrule
        ResNet18\cite{he2016deep} & 11.7 M & 1.8 B &   56.1$\%$    & 69.8\% \\
        DenseNet121\cite{huang2017densely} & \textbf{8.0 M} & 2.9 B &   29.3$\%$    & 74.6\% \\
        BN-Inception\cite{ioffe2015batch} & 11.2 M & 2.1 B &   48.8$\%$     & 73.5\% \\
        \ssd & 25.9 M & 1.9 B &     53.7$\%$  & 74.9\% \\
        \textbf{\css} & 13.5 M   &  \textbf{1.8 B}     &  \textbf{56.1$\%$}    & \textbf{75.6\%}  \\
        \bottomrule
        \end{tabular}%
    }
  \label{tab:backbones}%
\end{table}%

\begin{table}[tbp]
  \centering
  \caption{Results of ablation experiments on ImageNet-1K.}
    \begin{tabular}{l||r||r||r||r}
    \toprule
    \textbf{Method} & \textbf{\#Params} & \textbf{\#MACs} & \textbf{Latency} & \textbf{Acc@1} \\
    \midrule
    ResNet50 & 25.6 M & 4.1 B & 10.6 ms & 76.0\% \\
    \midrule
    \rcc   & 25.6 M & 1.1 B & 3.0 ms & 70.0\% \\
    RandomCrop & 25.6 M & 1.1 B & 2.9 ms & 69.1\% \\
    \textbf{\ssd} & 25.9 M & 1.2 B & 3.8 ms & \textbf{73.1\%} \\
    \midrule
    ResShrink & 25.6 M & 1.1 B & 3.0 ms & 70.0\% \\
    WidthShrink & 8.9 M & 1.2 B & 5.2 ms & 71.2\% \\
    DepthShrink & 11.1 M & 1.7 B & 5.0 ms & 69.6\% \\
    \textbf{\cs} & 11.4 M & 1.1 B & 4.6 ms & \textbf{71.5\%} \\
    \midrule
    \textbf{\css} &   11.7 M    &   1.3 B    &  5.5 ms  & \textbf{74.2\%}  \\
    \bottomrule
    \end{tabular}%
  \label{tab:ablation}%
\end{table}%

\subsection{Visualization}

We visualize the bounding boxes generated from both Grad-CAM and our predictor in Figure \ref{fig:visualbox}. We can see that the foreground of the most of images only occupies part of the whole images, thus performing inference on the whole image is unnecessary and inefficient, which coincides with our motivation. Moreover, Figure \ref{fig:visualbox} validates that, even though the foreground predictor only contains very limited computation and parameters for higher execution efficiency, it can also accurately localize the main object. 
With the accurate and quick prediction of the foreground, we are able to remove the background redundancy in images, thereby accelerating the execution of CNNs on resource-constrained edge devices.

\subsection{Ablation Study}

To validate the efficacy of both the DIC module and the CS module, we perform  comprehensive ablation experiments on ImageNet-1K. First, we only keep the DIC module to separately evaluate its efficacy. The results in Table \ref{tab:ablation} show that \ssd outperforms RCC and RandomCrop remarkably. Specifically, \ssd improves the top-1 accuracy by 3.1\% compared to RCC with a similar model cost. Subsequently, we evaluate the CS component by comparing it with single-dimension shrinking methods. The results also demonstrate the efficacy of the CS component. Finally, we combine the two modules into the completed framework \css. Because of the novel design of DIC and CS, we significantly improve the accuracy with a slightly higher latency on Xavier, optimizing the trade-off between accuracy and latency. The ablation experiments reveal that both DIC and CS contribute to the final performance.

\section{Conclusion}
\label{sec:conlusion}

In this paper, we propose a deep compression framework, \ssr, to comprehensively reduce the redundancy in both input images and network architectures, thereby facilitating the efficient inference on edge devices. First, we rethink the image cropping strategy for CNNs, and we find that existing static cropping methods introduce much redundancy in images. To reduce the redundancy in images, we propose a dynamic image cropping framework. The proposed framework first employs a foreground predictor with negligible overhead to quickly localize the important foreground region, and only the detected foreground will be utilized for inference, which remarkably reduces the computational cost. Meanwhile, to remove the redundancy in network architectures, we also propose a compound shrinking strategy, which coordinately shrinks the three dimensions of CNNs according to their importance to the final accuracy. By this means, the compound shrinking strategy is expected to comprehensively compress CNNs. Finally, the dynamic image cropping and compound shrinking are integrated to work coordinately for the optimal trade-off between accuracy and costs. Extensive and comprehensive experiments demonstrate the efficacy and efficiency of our approach.

\section{Acknowledgement}
This study is partially supported under the RIE2020 Industry Alignment Fund – Industry Collaboration Projects (IAF-ICP) Funding Initiative, as well as cash and in-kind contribution from the industry partner, HP Inc., through the HP-NTU Digital Manufacturing Corporate Lab (I1801E0028). This work is also partially supported by Nanyang Technological University, Singapore, under its NAP (M4082282).
\cleardoublepage
\bibliographystyle{ACM-Reference-Format}
\bibliography{acmart}


\begin{thebibliography}{47}

%%% the Web address.

\ifx \showCODEN    \undefined \def \showCODEN     #1{\unskip}     \fi
\ifx \showDOI      \undefined \def \showDOI       #1{#1}\fi
\ifx \showISBNx    \undefined \def \showISBNx     #1{\unskip}     \fi
\ifx \showISBNxiii \undefined \def \showISBNxiii  #1{\unskip}     \fi
\ifx \showISSN     \undefined \def \showISSN      #1{\unskip}     \fi
\ifx \showLCCN     \undefined \def \showLCCN      #1{\unskip}     \fi
\ifx \shownote     \undefined \def \shownote      #1{#1}          \fi
\ifx \showarticletitle \undefined \def \showarticletitle #1{#1}   \fi
\ifx \showURL      \undefined \def \showURL       {\relax}        \fi
% invisible to TeX
\providecommand\bibfield[2]{#2}
\providecommand\bibinfo[2]{#2}
\providecommand\natexlab[1]{#1}
\providecommand\showeprint[2][]{arXiv:#2}

\bibitem[Bochkovskiy et~al\mbox{.}(2020)]%
        {bochkovskiy2020yolov4}
\bibfield{author}{\bibinfo{person}{Alexey Bochkovskiy},
  \bibinfo{person}{Chien-Yao Wang}, {and} \bibinfo{person}{Hong-Yuan~Mark
  Liao}.} \bibinfo{year}{2020}\natexlab{}.
\newblock \showarticletitle{YOLOv4: Optimal Speed and Accuracy of Object
  Detection}.
\newblock \bibinfo{journal}{\emph{arXiv preprint arXiv:2004.10934}}
  (\bibinfo{year}{2020}).
\newblock


\bibitem[Brown et~al\mbox{.}(2020)]%
        {brown2020language}
\bibfield{author}{\bibinfo{person}{Tom Brown}, \bibinfo{person}{Benjamin Mann},
  \bibinfo{person}{Nick Ryder}, \bibinfo{person}{Melanie Subbiah},
  \bibinfo{person}{Jared~D Kaplan}, \bibinfo{person}{Prafulla Dhariwal},
  \bibinfo{person}{Arvind Neelakantan}, \bibinfo{person}{Pranav Shyam},
  \bibinfo{person}{Girish Sastry}, \bibinfo{person}{Amanda Askell},
  {et~al\mbox{.}}} \bibinfo{year}{2020}\natexlab{}.
\newblock \showarticletitle{Language Models Are Few-Shot Learners}. In
  \bibinfo{booktitle}{\emph{Advances in Neural Information Processing
  Systems}}, Vol.~\bibinfo{volume}{33}. \bibinfo{pages}{1877--1901}.
\newblock


\bibitem[Cordts et~al\mbox{.}(2016)]%
        {cordts2016cityscapes}
\bibfield{author}{\bibinfo{person}{Marius Cordts}, \bibinfo{person}{Mohamed
  Omran}, \bibinfo{person}{Sebastian Ramos}, \bibinfo{person}{Timo Rehfeld},
  \bibinfo{person}{Markus Enzweiler}, \bibinfo{person}{Rodrigo Benenson},
  \bibinfo{person}{Uwe Franke}, \bibinfo{person}{Stefan Roth}, {and}
  \bibinfo{person}{Bernt Schiele}.} \bibinfo{year}{2016}\natexlab{}.
\newblock \showarticletitle{The Cityscapes Dataset for Semantic Urban Scene
  Understanding}. In \bibinfo{booktitle}{\emph{Proceedings of the IEEE/CVF
  Conference on Computer Vision and Pattern Recognition}}.
  \bibinfo{pages}{3213--3223}.
\newblock


\bibitem[Deng et~al\mbox{.}(2009)]%
        {deng2009imagenet}
\bibfield{author}{\bibinfo{person}{Jia Deng}, \bibinfo{person}{Wei Dong},
  \bibinfo{person}{Richard Socher}, \bibinfo{person}{Li-Jia Li},
  \bibinfo{person}{Kai Li}, {and} \bibinfo{person}{Li Fei-Fei}.}
  \bibinfo{year}{2009}\natexlab{}.
\newblock \showarticletitle{ImageNet: A Large-Scale Hierarchical Image
  Database}. In \bibinfo{booktitle}{\emph{IEEE/CVF Conference on Computer
  Vision and Pattern Recognition}}. \bibinfo{pages}{248--255}.
\newblock


\bibitem[Ding et~al\mbox{.}(2019)]%
        {ding2019centripetal}
\bibfield{author}{\bibinfo{person}{Xiaohan Ding}, \bibinfo{person}{Guiguang
  Ding}, \bibinfo{person}{Yuchen Guo}, {and} \bibinfo{person}{Jungong Han}.}
  \bibinfo{year}{2019}\natexlab{}.
\newblock \showarticletitle{Centripetal SGD for Pruning Very Deep Convolutional
  Networks with Complicated Structure}. In
  \bibinfo{booktitle}{\emph{Proceedings of the IEEE/CVF Conference on Computer
  Vision and Pattern Recognition}}. \bibinfo{pages}{4943--4953}.
\newblock


\bibitem[Everingham et~al\mbox{.}(2010)]%
        {Everingham10}
\bibfield{author}{\bibinfo{person}{M. Everingham}, \bibinfo{person}{L.
  Van~Gool}, \bibinfo{person}{C.~K.~I. Williams}, \bibinfo{person}{J. Winn},
  {and} \bibinfo{person}{A. Zisserman}.} \bibinfo{year}{2010}\natexlab{}.
\newblock \showarticletitle{The Pascal Visual Object Classes (VOC) Challenge}.
\newblock \bibinfo{journal}{\emph{International Journal of Computer Vision}}
  \bibinfo{volume}{88}, \bibinfo{number}{2} (\bibinfo{date}{June}
  \bibinfo{year}{2010}), \bibinfo{pages}{303--338}.
\newblock


\bibitem[Fedus et~al\mbox{.}(2022)]%
        {fedus2022switch}
\bibfield{author}{\bibinfo{person}{William Fedus}, \bibinfo{person}{Barret
  Zoph}, {and} \bibinfo{person}{Noam Shazeer}.}
  \bibinfo{year}{2022}\natexlab{}.
\newblock \showarticletitle{Switch Transformers: Scaling to Trillion Parameter
  Models with Simple and Efficient Sparsity}.
\newblock \bibinfo{journal}{\emph{Journal of Machine Learning Research}}
  \bibinfo{volume}{23}, \bibinfo{number}{120} (\bibinfo{year}{2022}),
  \bibinfo{pages}{1--39}.
\newblock


\bibitem[Gao et~al\mbox{.}(2021)]%
        {gao2021network}
\bibfield{author}{\bibinfo{person}{Shangqian Gao}, \bibinfo{person}{Feihu
  Huang}, \bibinfo{person}{Weidong Cai}, {and} \bibinfo{person}{Heng Huang}.}
  \bibinfo{year}{2021}\natexlab{}.
\newblock \showarticletitle{Network Pruning via Performance Maximization}. In
  \bibinfo{booktitle}{\emph{Proceedings of the IEEE/CVF Conference on Computer
  Vision and Pattern Recognition}}. \bibinfo{pages}{9270--9280}.
\newblock


\bibitem[Han et~al\mbox{.}(2015)]%
        {han2015deep}
\bibfield{author}{\bibinfo{person}{Song Han}, \bibinfo{person}{Huizi Mao},
  {and} \bibinfo{person}{William~J Dally}.} \bibinfo{year}{2015}\natexlab{}.
\newblock \showarticletitle{Deep Compression: Compressing Deep Neural Networks
  with Pruning, Trained Quantization and Huffman Coding}.
\newblock \bibinfo{journal}{\emph{arXiv preprint arXiv:1510.00149}}
  (\bibinfo{year}{2015}).
\newblock


\bibitem[He et~al\mbox{.}(2017)]%
        {he2017mask}
\bibfield{author}{\bibinfo{person}{Kaiming He}, \bibinfo{person}{Georgia
  Gkioxari}, \bibinfo{person}{Piotr Doll{\'a}r}, {and} \bibinfo{person}{Ross
  Girshick}.} \bibinfo{year}{2017}\natexlab{}.
\newblock \showarticletitle{Mask R-CNN}. In
  \bibinfo{booktitle}{\emph{Proceedings of the IEEE/CVF International
  Conference on Computer Vision}}. \bibinfo{pages}{2961--2969}.
\newblock


\bibitem[He et~al\mbox{.}(2016)]%
        {he2016deep}
\bibfield{author}{\bibinfo{person}{Kaiming He}, \bibinfo{person}{Xiangyu
  Zhang}, \bibinfo{person}{Shaoqing Ren}, {and} \bibinfo{person}{Jian Sun}.}
  \bibinfo{year}{2016}\natexlab{}.
\newblock \showarticletitle{Deep Residual Learning for Image Recognition}. In
  \bibinfo{booktitle}{\emph{Proceedings of the IEEE/CVF Conference on Computer
  Vision and Pattern Recognition}}. \bibinfo{pages}{770--778}.
\newblock


\bibitem[Huang et~al\mbox{.}(2017)]%
        {huang2017densely}
\bibfield{author}{\bibinfo{person}{Gao Huang}, \bibinfo{person}{Zhuang Liu},
  \bibinfo{person}{Laurens Van Der~Maaten}, {and} \bibinfo{person}{Kilian~Q
  Weinberger}.} \bibinfo{year}{2017}\natexlab{}.
\newblock \showarticletitle{Densely Connected Convolutional Networks}. In
  \bibinfo{booktitle}{\emph{Proceedings of the IEEE/CVF Conference on Computer
  Vision and Pattern Recognition}}. \bibinfo{pages}{4700--4708}.
\newblock


\bibitem[Huang and Wang(2018)]%
        {huang2018data}
\bibfield{author}{\bibinfo{person}{Zehao Huang} {and} \bibinfo{person}{Naiyan
  Wang}.} \bibinfo{year}{2018}\natexlab{}.
\newblock \showarticletitle{Data-Driven Sparse Structure Selection for Deep
  Neural Networks}. In \bibinfo{booktitle}{\emph{Proceedings of the European
  Conference on Computer Vision}}. \bibinfo{pages}{304--320}.
\newblock


\bibitem[Ioffe and Szegedy(2015)]%
        {ioffe2015batch}
\bibfield{author}{\bibinfo{person}{Sergey Ioffe} {and}
  \bibinfo{person}{Christian Szegedy}.} \bibinfo{year}{2015}\natexlab{}.
\newblock \showarticletitle{Batch Normalization: Accelerating Deep Network
  Training by Reducing Internal Covariate Shift}. In
  \bibinfo{booktitle}{\emph{International Conference on Machine Learning}}.
  \bibinfo{pages}{448--456}.
\newblock


\bibitem[Kingma and Ba(2015)]%
        {kingma2015adam}
\bibfield{author}{\bibinfo{person}{Diederik~P Kingma} {and}
  \bibinfo{person}{Jimmy Ba}.} \bibinfo{year}{2015}\natexlab{}.
\newblock \showarticletitle{Adam: A Method for Stochastic Optimization}. In
  \bibinfo{booktitle}{\emph{International Conference on Learning
  Representations}}.
\newblock


\bibitem[Li et~al\mbox{.}(2018)]%
        {li2018tiny}
\bibfield{author}{\bibinfo{person}{Yuxi Li}, \bibinfo{person}{Jiuwei Li},
  \bibinfo{person}{Weiyao Lin}, {and} \bibinfo{person}{Jianguo Li}.}
  \bibinfo{year}{2018}\natexlab{}.
\newblock \showarticletitle{Tiny-DSOD: Lightweight Object Detection for
  Resource-Restricted Usages}. In \bibinfo{booktitle}{\emph{British Machine
  Vision Conference 2018}}.
\newblock


\bibitem[Li and Arora(2020)]%
        {Li2020An}
\bibfield{author}{\bibinfo{person}{Zhiyuan Li} {and} \bibinfo{person}{Sanjeev
  Arora}.} \bibinfo{year}{2020}\natexlab{}.
\newblock \showarticletitle{An Exponential Learning Rate Schedule for Deep
  Learning}. In \bibinfo{booktitle}{\emph{International Conference on Learning
  Representations}}.
\newblock


\bibitem[Liebenwein et~al\mbox{.}(2019)]%
        {liebenwein2019provable}
\bibfield{author}{\bibinfo{person}{Lucas Liebenwein}, \bibinfo{person}{Cenk
  Baykal}, \bibinfo{person}{Harry Lang}, \bibinfo{person}{Dan Feldman}, {and}
  \bibinfo{person}{Daniela Rus}.} \bibinfo{year}{2019}\natexlab{}.
\newblock \showarticletitle{Provable Filter Pruning for Efficient Neural
  Networks}. In \bibinfo{booktitle}{\emph{International Conference on Learning
  Representations}}.
\newblock


\bibitem[Lin et~al\mbox{.}(2020)]%
        {lin2020hrank}
\bibfield{author}{\bibinfo{person}{Mingbao Lin}, \bibinfo{person}{Rongrong Ji},
  \bibinfo{person}{Yan Wang}, \bibinfo{person}{Yichen Zhang},
  \bibinfo{person}{Baochang Zhang}, \bibinfo{person}{Yonghong Tian}, {and}
  \bibinfo{person}{Ling Shao}.} \bibinfo{year}{2020}\natexlab{}.
\newblock \showarticletitle{HRank: Filter Pruning Using High-Rank Feature Map}.
  In \bibinfo{booktitle}{\emph{Proceedings of the IEEE/CVF Conference on
  Computer Vision and Pattern Recognition}}. \bibinfo{pages}{1529--1538}.
\newblock


\bibitem[Lin et~al\mbox{.}(2019)]%
        {lin2019towards}
\bibfield{author}{\bibinfo{person}{Shaohui Lin}, \bibinfo{person}{Rongrong Ji},
  \bibinfo{person}{Chenqian Yan}, \bibinfo{person}{Baochang Zhang},
  \bibinfo{person}{Liujuan Cao}, \bibinfo{person}{Qixiang Ye},
  \bibinfo{person}{Feiyue Huang}, {and} \bibinfo{person}{David Doermann}.}
  \bibinfo{year}{2019}\natexlab{}.
\newblock \showarticletitle{Towards Optimal Structured CNN Pruning via
  Generative Adversarial Learning}. In \bibinfo{booktitle}{\emph{Proceedings of
  the IEEE/CVF Conference on Computer Vision and Pattern Recognition}}.
  \bibinfo{pages}{2790--2799}.
\newblock


\bibitem[Lin et~al\mbox{.}(2014)]%
        {lin2014microsoft}
\bibfield{author}{\bibinfo{person}{Tsung-Yi Lin}, \bibinfo{person}{Michael
  Maire}, \bibinfo{person}{Serge Belongie}, \bibinfo{person}{James Hays},
  \bibinfo{person}{Pietro Perona}, \bibinfo{person}{Deva Ramanan},
  \bibinfo{person}{Piotr Doll{\'a}r}, {and} \bibinfo{person}{C~Lawrence
  Zitnick}.} \bibinfo{year}{2014}\natexlab{}.
\newblock \showarticletitle{Microsoft COCO: Common Objects in Context}. In
  \bibinfo{booktitle}{\emph{European Conference on Computer Vision}}.
  \bibinfo{pages}{740--755}.
\newblock


\bibitem[Liu et~al\mbox{.}(2018)]%
        {liu2018darts}
\bibfield{author}{\bibinfo{person}{Hanxiao Liu}, \bibinfo{person}{Karen
  Simonyan}, {and} \bibinfo{person}{Yiming Yang}.}
  \bibinfo{year}{2018}\natexlab{}.
\newblock \showarticletitle{DARTS: Differentiable Architecture Search}. In
  \bibinfo{booktitle}{\emph{International Conference on Learning
  Representations}}.
\newblock


\bibitem[Liu et~al\mbox{.}(2016)]%
        {liu2016ssd}
\bibfield{author}{\bibinfo{person}{Wei Liu}, \bibinfo{person}{Dragomir
  Anguelov}, \bibinfo{person}{Dumitru Erhan}, \bibinfo{person}{Christian
  Szegedy}, \bibinfo{person}{Scott Reed}, \bibinfo{person}{Cheng-Yang Fu},
  {and} \bibinfo{person}{Alexander~C Berg}.} \bibinfo{year}{2016}\natexlab{}.
\newblock \showarticletitle{SSD: Single Shot Multibox Detector}. In
  \bibinfo{booktitle}{\emph{European Conference on Computer Vision}}.
  \bibinfo{pages}{21--37}.
\newblock


\bibitem[Liu et~al\mbox{.}(2022)]%
        {liu2022convnet}
\bibfield{author}{\bibinfo{person}{Zhuang Liu}, \bibinfo{person}{Hanzi Mao},
  \bibinfo{person}{Chao-Yuan Wu}, \bibinfo{person}{Christoph Feichtenhofer},
  \bibinfo{person}{Trevor Darrell}, {and} \bibinfo{person}{Saining Xie}.}
  \bibinfo{year}{2022}\natexlab{}.
\newblock \showarticletitle{A ConvNet for the 2020s}.
\newblock \bibinfo{journal}{\emph{Proceedings of the IEEE/CVF Conference on
  Computer Vision and Pattern Recognition}}.
\newblock


\bibitem[Molchanov et~al\mbox{.}(2019)]%
        {molchanov2019importance}
\bibfield{author}{\bibinfo{person}{Pavlo Molchanov}, \bibinfo{person}{Arun
  Mallya}, \bibinfo{person}{Stephen Tyree}, \bibinfo{person}{Iuri Frosio},
  {and} \bibinfo{person}{Jan Kautz}.} \bibinfo{year}{2019}\natexlab{}.
\newblock \showarticletitle{Importance Estimation for Neural Network Pruning}.
  In \bibinfo{booktitle}{\emph{Proceedings of the IEEE/CVF Conference on
  Computer Vision and Pattern Recognition}}. \bibinfo{pages}{11264--11272}.
\newblock


\bibitem[Radosavovic et~al\mbox{.}(2020)]%
        {radosavovic2020designing}
\bibfield{author}{\bibinfo{person}{Ilija Radosavovic},
  \bibinfo{person}{Raj~Prateek Kosaraju}, \bibinfo{person}{Ross Girshick},
  \bibinfo{person}{Kaiming He}, {and} \bibinfo{person}{Piotr Doll{\'a}r}.}
  \bibinfo{year}{2020}\natexlab{}.
\newblock \showarticletitle{Designing Network Design Spaces}. In
  \bibinfo{booktitle}{\emph{Proceedings of the IEEE/CVF Conference on Computer
  Vision and Pattern Recognition}}. \bibinfo{pages}{10428--10436}.
\newblock


\bibitem[Ren et~al\mbox{.}(2015)]%
        {ren2015faster}
\bibfield{author}{\bibinfo{person}{Shaoqing Ren}, \bibinfo{person}{Kaiming He},
  \bibinfo{person}{Ross Girshick}, {and} \bibinfo{person}{Jian Sun}.}
  \bibinfo{year}{2015}\natexlab{}.
\newblock \showarticletitle{Faster R-CNN: Towards Real-Time Object Detection
  with Region Proposal Networks}. In \bibinfo{booktitle}{\emph{Advances in
  Neural Information Processing Systems}}, Vol.~\bibinfo{volume}{28}.
\newblock


\bibitem[Ridnik et~al\mbox{.}(2021)]%
        {ridnik2021imagenet}
\bibfield{author}{\bibinfo{person}{Tal Ridnik}, \bibinfo{person}{Emanuel
  Ben-Baruch}, \bibinfo{person}{Asaf Noy}, {and} \bibinfo{person}{Lihi
  Zelnik-Manor}.} \bibinfo{year}{2021}\natexlab{}.
\newblock \showarticletitle{ImageNet-21K Pretraining for the Masses}. In
  \bibinfo{booktitle}{\emph{Advances in Neural Information Processing
  Systems}}, Vol.~\bibinfo{volume}{34}.
\newblock


\bibitem[Sandler et~al\mbox{.}(2018)]%
        {sandler2018mobilenetv2}
\bibfield{author}{\bibinfo{person}{Mark Sandler}, \bibinfo{person}{Andrew
  Howard}, \bibinfo{person}{Menglong Zhu}, \bibinfo{person}{Andrey Zhmoginov},
  {and} \bibinfo{person}{Liang-Chieh Chen}.} \bibinfo{year}{2018}\natexlab{}.
\newblock \showarticletitle{MobileNetv2: Inverted Residuals and Linear
  Bottlenecks}. In \bibinfo{booktitle}{\emph{Proceedings of the IEEE/CVF
  Conference on Computer Vision and Pattern Recognition}}.
  \bibinfo{pages}{4510--4520}.
\newblock


\bibitem[Selvaraju et~al\mbox{.}(2017)]%
        {selvaraju2017grad}
\bibfield{author}{\bibinfo{person}{Ramprasaath~R Selvaraju},
  \bibinfo{person}{Michael Cogswell}, \bibinfo{person}{Abhishek Das},
  \bibinfo{person}{Ramakrishna Vedantam}, \bibinfo{person}{Devi Parikh}, {and}
  \bibinfo{person}{Dhruv Batra}.} \bibinfo{year}{2017}\natexlab{}.
\newblock \showarticletitle{Grad-CAM: Visual Explanations from Deep Networks
  via Gradient-Based Localization}. In \bibinfo{booktitle}{\emph{Proceedings of
  the IEEE/CVF International Conference on Computer Vision}}.
  \bibinfo{pages}{618--626}.
\newblock


\bibitem[Shi et~al\mbox{.}(2016)]%
        {shi2016edge}
\bibfield{author}{\bibinfo{person}{Weisong Shi}, \bibinfo{person}{Jie Cao},
  \bibinfo{person}{Quan Zhang}, \bibinfo{person}{Youhuizi Li}, {and}
  \bibinfo{person}{Lanyu Xu}.} \bibinfo{year}{2016}\natexlab{}.
\newblock \showarticletitle{Edge Computing: Vision and Challenges}.
\newblock \bibinfo{journal}{\emph{IEEE Internet of Things Journal}}
  \bibinfo{volume}{3} (\bibinfo{year}{2016}), \bibinfo{pages}{637--646}.
\newblock


\bibitem[Szegedy et~al\mbox{.}(2016)]%
        {szegedy2016rethinking}
\bibfield{author}{\bibinfo{person}{Christian Szegedy}, \bibinfo{person}{Vincent
  Vanhoucke}, \bibinfo{person}{Sergey Ioffe}, \bibinfo{person}{Jon Shlens},
  {and} \bibinfo{person}{Zbigniew Wojna}.} \bibinfo{year}{2016}\natexlab{}.
\newblock \showarticletitle{Rethinking The Inception Architecture for Computer
  Vision}. In \bibinfo{booktitle}{\emph{Proceedings of the IEEE/CVF Conference
  on Computer Vision and Pattern Recognition}}. \bibinfo{pages}{2818--2826}.
\newblock


\bibitem[Tan et~al\mbox{.}(2019)]%
        {tan2019mnasnet}
\bibfield{author}{\bibinfo{person}{Mingxing Tan}, \bibinfo{person}{Bo Chen},
  \bibinfo{person}{Ruoming Pang}, \bibinfo{person}{Vijay Vasudevan},
  \bibinfo{person}{Mark Sandler}, \bibinfo{person}{Andrew Howard}, {and}
  \bibinfo{person}{Quoc~V Le}.} \bibinfo{year}{2019}\natexlab{}.
\newblock \showarticletitle{MnasNet: Platform-Aware Neural Architecture Search
  for Mobile}. In \bibinfo{booktitle}{\emph{Proceedings of the IEEE/CVF
  Conference on Computer Vision and Pattern Recognition}}.
  \bibinfo{pages}{2820--2828}.
\newblock


\bibitem[Tan and Le(2019)]%
        {tan2019efficientnet}
\bibfield{author}{\bibinfo{person}{Mingxing Tan} {and} \bibinfo{person}{Quoc
  Le}.} \bibinfo{year}{2019}\natexlab{}.
\newblock \showarticletitle{Efficientnet: Rethinking Model Scaling for
  Convolutional Neural Networks}. In \bibinfo{booktitle}{\emph{International
  Conference on Machine Learning}}. \bibinfo{pages}{6105--6114}.
\newblock


\bibitem[Wang et~al\mbox{.}(2020)]%
        {wang2020glance}
\bibfield{author}{\bibinfo{person}{Yulin Wang}, \bibinfo{person}{Kangchen Lv},
  \bibinfo{person}{Rui Huang}, \bibinfo{person}{Shiji Song},
  \bibinfo{person}{Le Yang}, {and} \bibinfo{person}{Gao Huang}.}
  \bibinfo{year}{2020}\natexlab{}.
\newblock \showarticletitle{Glance and Focus: A Dynamic Approach to Reducing
  Spatial Redundancy in Image Classification}. In
  \bibinfo{booktitle}{\emph{Advances in Neural Information Processing
  Systems}}, Vol.~\bibinfo{volume}{33}. \bibinfo{pages}{2432--2444}.
\newblock


\bibitem[Wang et~al\mbox{.}(2018)]%
        {wang2018learning}
\bibfield{author}{\bibinfo{person}{Yunhe Wang}, \bibinfo{person}{Chang Xu},
  \bibinfo{person}{Chunjing Xu}, \bibinfo{person}{Chao Xu}, {and}
  \bibinfo{person}{Dacheng Tao}.} \bibinfo{year}{2018}\natexlab{}.
\newblock \showarticletitle{Learning Versatile Filters for Efficient
  Convolutional Neural Networks}. In \bibinfo{booktitle}{\emph{Advances in
  Neural Information Processing Systems}}, Vol.~\bibinfo{volume}{31}.
\newblock


\bibitem[Wang et~al\mbox{.}(2021)]%
        {wang2021convolutional}
\bibfield{author}{\bibinfo{person}{Zi Wang}, \bibinfo{person}{Chengcheng Li},
  {and} \bibinfo{person}{Xiangyang Wang}.} \bibinfo{year}{2021}\natexlab{}.
\newblock \showarticletitle{Convolutional Neural Network Pruning with
  Structural Redundancy Reduction}. In \bibinfo{booktitle}{\emph{Proceedings of
  the IEEE/CVF Conference on Computer Vision and Pattern Recognition}}.
  \bibinfo{pages}{14913--14922}.
\newblock


\bibitem[Wei et~al\mbox{.}(2019)]%
        {wei2019unsupervised}
\bibfield{author}{\bibinfo{person}{Xiu-Shen Wei}, \bibinfo{person}{Chen-Lin
  Zhang}, \bibinfo{person}{Jianxin Wu}, \bibinfo{person}{Chunhua Shen}, {and}
  \bibinfo{person}{Zhi-Hua Zhou}.} \bibinfo{year}{2019}\natexlab{}.
\newblock \showarticletitle{Unsupervised Object Discovery and Co-Localization
  by Deep Descriptor Transformation}.
\newblock \bibinfo{journal}{\emph{Pattern Recognition}}  \bibinfo{volume}{88}
  (\bibinfo{year}{2019}), \bibinfo{pages}{113--126}.
\newblock


\bibitem[Wu et~al\mbox{.}(2019)]%
        {wu2019fbnet}
\bibfield{author}{\bibinfo{person}{Bichen Wu}, \bibinfo{person}{Xiaoliang Dai},
  \bibinfo{person}{Peizhao Zhang}, \bibinfo{person}{Yanghan Wang},
  \bibinfo{person}{Fei Sun}, \bibinfo{person}{Yiming Wu},
  \bibinfo{person}{Yuandong Tian}, \bibinfo{person}{Peter Vajda},
  \bibinfo{person}{Yangqing Jia}, {and} \bibinfo{person}{Kurt Keutzer}.}
  \bibinfo{year}{2019}\natexlab{}.
\newblock \showarticletitle{FBnet: Hardware-Aware Efficient Convnet Design via
  Differentiable Neural Architecture Search}. In
  \bibinfo{booktitle}{\emph{Proceedings of the IEEE/CVF Conference on Computer
  Vision and Pattern Recognition}}. \bibinfo{pages}{10734--10742}.
\newblock


\bibitem[Yang et~al\mbox{.}(2020)]%
        {yang2020resolution}
\bibfield{author}{\bibinfo{person}{Le Yang}, \bibinfo{person}{Yizeng Han},
  \bibinfo{person}{Xi Chen}, \bibinfo{person}{Shiji Song},
  \bibinfo{person}{Jifeng Dai}, {and} \bibinfo{person}{Gao Huang}.}
  \bibinfo{year}{2020}\natexlab{}.
\newblock \showarticletitle{Resolution Adaptive Networks for Efficient
  Inference}. In \bibinfo{booktitle}{\emph{Proceedings of the IEEE/CVF
  Conference on Computer Vision and Pattern Recognition}}.
  \bibinfo{pages}{2369--2378}.
\newblock


\bibitem[Yu et~al\mbox{.}(2019)]%
        {yu2019slimmable}
\bibfield{author}{\bibinfo{person}{Jiahui Yu}, \bibinfo{person}{Linjie Yang},
  \bibinfo{person}{Ning Xu}, \bibinfo{person}{Jianchao Yang}, {and}
  \bibinfo{person}{Thomas Huang}.} \bibinfo{year}{2019}\natexlab{}.
\newblock \showarticletitle{Slimmable Neural Networks}. In
  \bibinfo{booktitle}{\emph{International Conference on Learning
  Representations}}.
\newblock


\bibitem[Yun et~al\mbox{.}(2019)]%
        {yun2019cutmix}
\bibfield{author}{\bibinfo{person}{Sangdoo Yun}, \bibinfo{person}{Dongyoon
  Han}, \bibinfo{person}{Seong~Joon Oh}, \bibinfo{person}{Sanghyuk Chun},
  \bibinfo{person}{Junsuk Choe}, {and} \bibinfo{person}{Youngjoon Yoo}.}
  \bibinfo{year}{2019}\natexlab{}.
\newblock \showarticletitle{CutMix: Regularization Strategy to Train Strong
  Classifiers with Localizable Features}. In
  \bibinfo{booktitle}{\emph{Proceedings of the IEEE/CVF International
  Conference on Computer Vision}}. \bibinfo{pages}{6023--6032}.
\newblock


\bibitem[Zagoruyko and Komodakis(2016)]%
        {zagoruyko2016wide}
\bibfield{author}{\bibinfo{person}{Sergey Zagoruyko} {and}
  \bibinfo{person}{Nikos Komodakis}.} \bibinfo{year}{2016}\natexlab{}.
\newblock \showarticletitle{Wide Residual Networks}. In
  \bibinfo{booktitle}{\emph{British Machine Vision Conference 2016}}.
\newblock


\bibitem[Zhang et~al\mbox{.}(2020)]%
        {zhang2020rethinking}
\bibfield{author}{\bibinfo{person}{Chen-Lin Zhang}, \bibinfo{person}{Yun-Hao
  Cao}, {and} \bibinfo{person}{Jianxin Wu}.} \bibinfo{year}{2020}\natexlab{}.
\newblock \showarticletitle{Rethinking the Route Towards Weakly Supervised
  Object Localization}. In \bibinfo{booktitle}{\emph{Proceedings of the
  IEEE/CVF Conference on Computer Vision and Pattern Recognition}}.
  \bibinfo{pages}{13460--13469}.
\newblock


\bibitem[Zhang et~al\mbox{.}(2018)]%
        {zhang2018adversarial}
\bibfield{author}{\bibinfo{person}{Xiaolin Zhang}, \bibinfo{person}{Yunchao
  Wei}, \bibinfo{person}{Jiashi Feng}, \bibinfo{person}{Yi Yang}, {and}
  \bibinfo{person}{Thomas~S Huang}.} \bibinfo{year}{2018}\natexlab{}.
\newblock \showarticletitle{Adversarial Complementary Learning for Weakly
  Supervised Object Localization}. In \bibinfo{booktitle}{\emph{Proceedings of
  the IEEE/CVF Conference on Computer Vision and Pattern Recognition}}.
  \bibinfo{pages}{1325--1334}.
\newblock


\bibitem[Zhou et~al\mbox{.}(2016)]%
        {zhou2016learning}
\bibfield{author}{\bibinfo{person}{Bolei Zhou}, \bibinfo{person}{Aditya
  Khosla}, \bibinfo{person}{Agata Lapedriza}, \bibinfo{person}{Aude Oliva},
  {and} \bibinfo{person}{Antonio Torralba}.} \bibinfo{year}{2016}\natexlab{}.
\newblock \showarticletitle{Learning Deep Features for Discriminative
  Localization}. In \bibinfo{booktitle}{\emph{Proceedings of the IEEE/CVF
  Conference on Computer Vision and Pattern Recognition}}.
  \bibinfo{pages}{2921--2929}.
\newblock


\bibitem[Zhu et~al\mbox{.}(2021)]%
        {zhu2021dynamic}
\bibfield{author}{\bibinfo{person}{Mingjian Zhu}, \bibinfo{person}{Kai Han},
  \bibinfo{person}{Enhua Wu}, \bibinfo{person}{Qiulin Zhang},
  \bibinfo{person}{Ying Nie}, \bibinfo{person}{Zhenzhong Lan}, {and}
  \bibinfo{person}{Yunhe Wang}.} \bibinfo{year}{2021}\natexlab{}.
\newblock \showarticletitle{Dynamic Resolution Network}. In
  \bibinfo{booktitle}{\emph{Advances in Neural Information Processing
  Systems}}, Vol.~\bibinfo{volume}{34}.
\newblock


\end{thebibliography}

\appendix









\end{document}